\pdfoutput=1
\documentclass[10pt,a4paper]{article}
\usepackage[margin=25mm]{geometry}
\usepackage[T1]{fontenc}
\usepackage[utf8]{inputenc}
\usepackage{lmodern}
\usepackage{authblk}
\usepackage{microtype}

\usepackage[numbers]{natbib}

\usepackage{graphicx}
\usepackage{adjustbox}
\usepackage{booktabs}
\usepackage{multirow}
\usepackage{amsmath,amssymb,amsthm}

\usepackage{xcolor}
\usepackage{tikz}
\usetikzlibrary{positioning,arrows.meta,calc,shapes.geometric,backgrounds,fit}
\usepackage{tabularx}
\usepackage{url}

\usepackage{placeins}

\newcommand{\revx}[1]{{\color{black}#1}}
\newenvironment{revblock}{\color{black}}{}

\begin{document}

\title{Leakage-Aware Calibrated Decision-Level Fusion for Mental-Health Text Classification}
\author[1]{Wadii Boulila\thanks{Corresponding author: \texttt{wboulila@psu.edu.sa}}}
\author[2]{Giuseppe Varone}
\author[1]{Bilel Benjdira}
\author[3]{Maha Driss}
\affil[1]{Robotics and Internet-of-Things Laboratory, Prince Sultan University, Riyadh 11586, Saudi Arabia}
\affil[2]{Department of Psychiatry, Harvard Medical School, Boston, MA 02115, United States}
\affil[3]{College of Engineering and Advanced Computing, Alfaisal University, Riyadh 11533, Saudi Arabia}
\date{}
\maketitle

\begin{abstract}
Decision-level fusion improves transformer-based text categorization, but successful fusion requires model error compensation, calibrated ensemble probabilities, and data-leakage-free evaluations. We developed a leakage- and calibration-conscious method for heterogeneous transformer fusion and evaluated it using $262{,}276$ proxy-labeled Reddit entries about anxiety, depression, schizophrenia, and suicidal ideation. In a classic stratified split, MinHash--LSH identified $29.1\%$ of test entries as lexical near-duplicates of the training dataset, and augmentation-lineage reconstruction showed $50.2\%$ dependence, with $30.0\%$ of records retained after lexical filtering. Thus, analyses on lineage-disjoint partitions have been performed. Twelve encoders, three seeds of Bio-ClinicalBERT, MentalBERT, MentalRoBERTa, and RoBERTa-base, were carefully calibrated, and calibration and ensemble selection were done in cross-fitted validation. A class-specific Complementarity Index (CI), the Jaccard distance between the sets of validation examples each encoder classifies correctly, was computed from member predictions before aggregation and evaluated as a predictor of fusion gain. CI was positively correlated with independent fusion gain for depression ($r=0.50$) and suicidal ideation ($r=0.68$) but was not associated with anxiety and had mixed results for the minor schizophrenia category. The selected ensemble of MentalBERT, MentalRoBERTa, and RoBERTa-base achieved a macro-F1 of $0.844$ $[0.823,0.863]$, MCC of $0.779$, and ECE of $0.0033$ during post-fusion recalibration, improving macro-F1 by $0.021$ $[0.007,0.037]$ over the selected validation best single encoder. The variability of Integrated-Gradients attribution was $2.3$ times greater in fusion-rescued instances; however, intentionally omitting terms that drive divergence reduced rescues by $58\%$, compared with $12\%$ under random omission. Dependencies between training and evaluation records that were missed by lexical similarity alone were identified through augmentation-lineage reconstruction. Moreover, the analysis of class-specific errors and model calibration after fusion gives hints about the interactions between different transformer architectures, which are completely missed by overall accuracy metrics.
\end{abstract}

\noindent\textbf{Keywords:} Decision-Level Fusion; Transformer Ensembles; Complementarity Index; Model Calibration; Leakage-Aware Evaluation; Mental-Health Text Classification; Explainable AI

\section{Introduction}
\label{sec:intro}

Classifying mental-health-related text from social media remains a challenging supervised learning task. Public posts from self-selected mental-health communities on Reddit provide large volumes of labeled text that would be difficult to obtain through clinical recruitment. However, these labels are weak because posts are generally labeled according to the subreddit in which they appear. Therefore, the subreddit serves only as a proxy for a mental-health condition rather than as evidence of a clinically validated diagnosis. Using subreddits as proxy labels introduces three confounding factors that directly affect model evaluation. First, self-selection creates sampling bias because the data mainly represent users who are willing to discuss their mental health publicly~\citep{harrigian2021state}. Second, the labels may be noisy because users can post in a mental-health subreddit without meeting the formal diagnostic criteria for the associated condition. Third, many conditions share similar linguistic features. For example, depression, anxiety, and suicidal ideation may involve common affective, cognitive, and self-referential expressions~\citep{rude2004language,joiner2010interpersonal}. Therefore, a single-label classifier may learn dataset-specific patterns rather than reliable indicators of a particular condition, making closely related classes difficult to distinguish.

Obtaining clinically confirmed labels at a comparable scale is generally impractical because of privacy, consent, and data-access constraints. Despite their limitations, large text corpora have supported the development of domain-adapted language models. Bio-ClinicalBERT was pretrained on clinical notes~\citep{alsentzer2019clinicalbert}, whereas MentalBERT was adapted specifically to mental-health-related language~\citep{ji2022mentalbert}. This setting therefore provides a demanding testbed for assessing whether decision-level fusion can benefit from complementary model errors while maintaining well-calibrated confidence estimates.

Three limitations in current ensemble practice reduce the reliability of fusion in this setting. First, ensembles are most effective when their members make complementary errors. However, diversity is usually measured after training using global statistics, such as disagreement, correlation, or the double-fault measure~\citep{kuncheva2003measures}. These measures summarize overall diversity but do not identify beforehand which encoder combinations can improve performance or which classes will benefit from fusion. Second, averaging the outputs of individually calibrated models does not guarantee that the resulting ensemble will also be calibrated. Nevertheless, ensemble confidence scores are often treated as reliable without directly evaluating or calibrating the final output. Third, social-media datasets may contain duplicate or near-duplicate posts across the training and test sets. Such contamination can inflate performance and provide a misleading view of generalization. A reliable fusion method for noisy, proxy-labeled text should therefore address all three limitations: estimate the benefit of fusion before aggregation, calibrate the final ensemble output, and confirm that the evaluation splits are free from data leakage.

In this paper, these limitations have been addressed within a single pipeline. The system operates as decision-level fusion within the standard feature-, score-, and decision-level taxonomy of information fusion~\citep{khaleghi2013multisensor}. Each base classifier produces a calibrated probability
vector, and the fusion mechanism is the combination of those vectors at
the decision layer. The proposed pipeline consists of three components. First, \revx{a split-integrity audit combines MinHash--Locality-Sensitive Hashing to detect near-duplicate posts with a lineage audit that links each augmented record to its original post. The data are then split at the lineage level, ensuring that an original post and its derived versions cannot appear in different training and evaluation sets.} Second, a class-specific complementarity index (CI) is introduced as a pre-fusion criterion. Using predictions from the validation set, the CI estimates how well candidate encoders make complementary correct predictions for each class. Third, a two-stage calibration procedure first applies temperature scaling to each ensemble member and then recalibrates the fused output. Together, these components address three important concerns in decision-level fusion: data leakage, limited transparency in ensemble selection, and unreliable confidence estimates. The resulting framework supports accurate and better-calibrated predictions while providing a clearer basis for selecting the models included in the ensemble.

\begin{revblock}
The evaluation uses $262{,}276$ records from four proxy mental-health categories: anxiety, depression, schizophrenia, and suicidal ideation. The corpus includes $149{,}195$ original posts and $113{,}081$ augmented records. A stratified random split shows substantial contamination: MinHash--LSH flags $29.1\%$ of the test set. At the same time, the lineage audit reveals that $50.2\%$ of the test records share an augmentation lineage group with a training record, either as derivatives of training posts or as originals whose derivatives lie in training. This includes $30.0\%$ of the records that the lexical audit retains. Therefore, a lineage-disjoint split is constructed in which the validation and test sets contain only original posts.

Twelve encoders (three seeds from each of four model families) are fine-tuned. All model selection, calibration, and fusion decisions are made on the validation data under a cross-fitted protocol; the test set is reserved for final reporting. The selected ensemble combines MentalBERT, MentalRoBERTa, and RoBERTa-base and is recalibrated using $T^{\star}=0.886$. It achieves a macro-F1 of $0.844$ $[0.823,0.863]$, an MCC of $0.779$, and an ECE of $0.0033$. It improves macro-F1 by $0.021$ over the validation-selected best single encoder, with significant permutation and multiplicity-corrected McNemar tests.

Dependence-aware analyses show that CI predicts fusion gains for depression and suicidal ideation, but not for anxiety. The ensemble also transfers zero-shot to an author-disjoint binary dataset from Reddit. In addition, integrated Gradients and counterfactual word removal show how disagreement among the models supports successful fusion. Although evaluated on mental-health text, the framework can be applied to other multi-encoder classification tasks requiring leakage-aware evaluation, calibrated confidence, and transparent ensemble selection.
\end{revblock}

\paragraph{Contributions.} In this paper, we present the following contributions:
\begin{enumerate}
  \item \revx{\textbf{A leakage-aware decision-level fusion framework} is proposed for proxy-labeled mental-health text classification. The framework integrates a two-layer split-integrity audit (lexical near-duplicate detection and augmentation-lineage tracing), validation-driven encoder selection with a class-specific complementarity diagnostic, and calibration-aware probability aggregation within a unified pipeline. Applied to a stratified random split of the corpus, the lineage audit shows that half of the test records share an augmentation lineage group with a training record and that lexical filtering alone leaves $30.0\%$ of the retained records so linked; the lineage-disjoint split removes augmentation-lineage contamination and reduces the lexical near-duplicate rate to $0.4\%$ and the semantic (cosine $\geq0.90$) near-duplicate rate to $0.8\%$; contamination through shared authors, threads, or time could not be audited and is stated as an open limitation.}

  \item \revx{\textbf{A class-specific Complementarity Index (CI)}, equal to the Jaccard distance between the sets of class-$c$ validation examples each encoder classifies correctly, is introduced as a validation-computable predictor of where fusion will help, and is evaluated against test-set fusion gain with statistics that respect the dependence among pair--class observations (cluster bootstraps over pairs and over encoders, a mixed-effects model with class effects and a symmetric multi-membership random intercept per encoder, encoder-level bootstrap intervals within each class, and leave-one-encoder-out prediction). The association is class-dependent: significant within depression ($r=0.50$) and suicidal ideation ($r=0.68$), absent for anxiety, and uninformative for a $75$-record class.}

  \item \textbf{A two-stage calibration strategy} is developed for decision-level fusion. Individual encoders are first temperature-scaled separately, and the aggregated ensemble output is then recalibrated post hoc, improving probabilistic reliability without sacrificing predictive performance. \revx{The direction of the post-hoc correction is shown to be domain-dependent: the fused output of individually calibrated members is under-confident in domain ($T^{\star}<1$ in every validation fold) and over-confident under external transfer ($T^{\star}>1$).}

  \item \revx{\textbf{A compact heterogeneous three-member ensemble} composed of MentalBERT, MentalRoBERTa, and RoBERTa-base, selected under a cross-fitted validation protocol, achieves macro-F1 $=0.844$ $[0.823,0.863]$, MCC $=0.779$, and ECE $=0.0033$ on the lineage-disjoint test set, exceeding the validation-selected best single encoder by $+0.021$ macro-F1 (paired bootstrap $[+0.007,+0.037]$; permutation $p=0.0035$; McNemar $p=4.6\times10^{-4}$ after Holm correction over the $67$ comparisons performed); the $75$-post schizophrenia class contributes $71\%$ of that gain, and the improvement over the three larger classes alone is $+0.008$ $[+0.003,+0.014]$ and coming within $0.004$ of a twelve-model ensemble with one quarter as many forward passes per prediction (three encoders of the same size instead of twelve; wall-clock inference was not measured). Seed-equalized results (three seeds per family), the post-selection spread of the cross-fitted protocol ($0.842\pm0.004$), and zero-shot transfer to an author-disjoint, same-platform, two-class external corpus are reported.}

  \item \revx{\textbf{An attribution-diversity analysis with an interventional test} is conducted on the deployed checkpoints. Integrated-Gradients attributions are aligned at word level so that WordPiece and byte-pair tokenizers can be compared; fusion-rescued cases show $2.3\times$ higher inter-model attribution diversity than all-correct cases, the effect persists between member pairs that predict the same class, and removing the five words on which the correct members diverge from the incorrect member destroys $58\%$ of rescues versus $12\%$ for random words.}
\end{enumerate}

The remainder of the paper is organized as follows. Section~\ref{sec:related}
surveys related work with a comparative table.
Section~\ref{sec:theory} formalizes the theoretical foundations, including the
CI and the post-hoc calibration of aggregated predictions.
Section~\ref{sec:method} describes the \revx{corpus, split-integrity audit, lineage-disjoint split, fine-tuning setup, and evaluation protocol}.
Section~\ref{sec:results} presents the results and statistical tests.
Section~\ref{sec:xai} presents the \revx{attribution-diversity analysis and the counterfactual intervention}.
Section~\ref{sec:discussion} discusses implications and limitations.
Section~\ref{sec:conclusion} concludes the paper.

\section{Related Work}
\label{sec:related}
\revx{Recent work in mental-health natural language processing has moved from domain-adapted language models to heterogeneous transformer ensembles and strategies for evaluating their predictions. However, greater performance with model fusion does not prove that the constituent models provide complementary information, that aggregated probabilities remain properly calibrated, or that apparent improvements hold under stringent restrictions on data dependence. These challenges are especially relevant for social-media datasets with augmented or closely related observations. We cover four aspects directly related to reliable decision-level fusion: domain-adapted language models, classifier complementarity and ensemble learning, probability calibration and leakage-aware evaluation, and explanation of ensemble complementarity.

Pretraining on domain-specific data is necessary for transformer-based mental-health NLP. Alsentzer et al.~\citep{alsentzer2019clinicalbert} created variants of ClinicalBERT trained on clinical notes and showed that adapting to clinical language can improve performance on medical NLP tasks. Clinical notes, however, differ greatly from self-authored social-media posts in terms of terminology, writing style, and communicative context. Ji et al.~\citep{ji2022mentalbert} tackled this distinction more directly with MentalBERT and MentalRoBERTa, adapting them to mental-health-related social-media text. Their results suggest that further pretraining on language closer to the target domain can boost downstream mental-health NLP performance. Mezzi et al. ~\citep{mezzi2022mental} applied BERT to mental-health intent recognition for Arabic-speaking patients using MINI-based interview data collected at the Military Hospital of Tunis. More recently, S\'anchez Rodr\'iguez et al.~\citep{sanchez2026understanding} compared BERT and MentalBERT to classify Reddit posts from subreddits for anxiety, depression, bipolar disorder, and borderline personality disorder. MentalBERT showed a slight but persistent advantage over BERT, and Local Interpretable Model-agnostic Explanations (LIME) were used to identify language aspects associated with the predictions. These experiments show the importance of domain-adapted representations, but differences in encoder performance do not prove whether diverse representations provide complementary information that could be used through fusion.
There is a large body of work studying the relationship between classifier diversity and ensemble performance. Measures such as disagreement, correlation, Q-statistic, and double fault were studied by Shipp and Kuncheva~\citep{shipp2002relationships} and Kuncheva and Whitaker~\citep{kuncheva2003measures}. Brown et al.~\citep{brown2005diversity} also stressed that diversity is only relevant to prediction accuracy: a disagreement among classifiers does not necessarily mean their ensemble will outperform the individual models. However, heterogeneous transformer ensembles have recently been shown to be promising for mental-health text analysis. Kasap et al. \citep{kasap2026ensemble} applied DistilBERT, RoBERTa, Mental-BERT, Mental-RoBERTa, and DeBERTa for multi-label categorization of depression-related emotions and experimented with stacking, hard-voting, and soft-voting. Islam et al.~\citep{islam2026ensemble} proposed a heterogeneous transformer ensemble for depression related emotion and severity classification based on soft voting weighted by validation performance and post-hoc LIME explanations. The models used for the ensemble are based on DistilBERT, ALBERT, XLM-RoBERTa, and DeBERTa. These experiments support heterogeneous transformer fusion in mental-health NLP, but leave open a more specific question: can differences between candidate classifiers reveal combinations that are likely to benefit from fusion? 
Common classifier-diversity metrics characterize relationships among model outputs, but they do not directly measure class-specific non-overlap among correctly classified validation examples. This leads us to separate generic classifier diversity from correctness complementarity and check whether such complementarity is related to subsequent fusion gain.

Predictive accuracy and probabilistic trustworthiness are separate features of classifiers. Niculescu-Mizil and Caruana \citep{niculescu2005predicting} showed that strong classifiers are not necessarily reliable probability estimators. Guo et al. \citep{guo2017calibration} showed that neural networks can be severely poorly calibrated even with high classification accuracy and that temperature scaling is a simple and effective post-hoc calibration method. Calibration is especially important for decision-level fusion since probability-based ensembles act directly on the confidence estimates generated by their members. Averaging calibrated member probabilities does not necessarily produce a calibrated ensemble. The fused output therefore requires separate evaluation and, when needed, recalibration. Islam et al.~\citep{islam2026enhanced} used validation performance to guide Bayesian probability fusion for structured clinical data, while Demirta\c{s} Alpsalaz et al.~\citep{alpsalaz2026explainability} combined calibrated confidence with explainability-related information to get sample-specific ensemble weights. Abdar et al.~\citep{abdar2021review} also highlighted the larger significance of trustworthy uncertainty estimation in deep learning. Taken together, these findings support treating calibration of the fused predictive distribution as a separate challenge from calibrating the individual classifiers.

Evaluation also requires separation between training and evaluation observations. This is especially relevant when dealing with social-media information collected from a variety of public sources, because exact, near-duplicate, reposted, or altered versions of the same underlying post can be present in nominally distinct partitions. Harrigian et al.~\citep{harrigian2021state} identified dataset development, sampling, labeling, and evaluation as important methodological challenges in applying NLP to mental health, and Chancellor and De Choudhury~\citep{chancellor2020methods} pointed out the relevance of data and methodological decisions for the interpretation of social-media-based mental-health predictions. Lee et al. \citep{lee2022deduplicating} demonstrated that repeated training instances can encourage memorization and drive explicit deduplication and overlap auditing. But lexical similarity is one type of dependence. An augmented post can differ enough from its source to fall below a lexical-similarity criterion and still derive from the same underlying observation.

The effects of leaking from preprocessing have been recently proved directly in mental-health social-media classification. Hammad and Rababah~\citep{hammad2026domain} showed that applying segmentation and oversampling before the train–test split could include instances generated from the same original post being included in both the training and evaluation collections. The authors also replaced this technique with an original-post-level split and used oversampling only on training data, which lowered their reported macro-F1 from $\sim 0.96$ to $0.67$. We see that dependencies imposed before partitioning can considerably exaggerate perceived generalization performance. More generally, lexical deduplication alone may be insufficient to detect all relationships induced by data augmentation, since a modified text may be lexically distinct yet from the same underlying observation. This divergence justifies explicit treatment of augmentation lineage in addition to traditional near-duplicate detection in the evaluation of models on enhanced social-media corpora.

Explainable-AI approaches provide an additional view on model complementarity, as classifiers may reach the same or different results depending on different parts of the input. Integrated Gradients~\citep{sundararajan2017ig} gives attributions of the output of a differentiable model to the input features, and SHAP~\citep{lundberg2017unified} provides feature-level explanations based on Shapley values. S\'anchez Rodr\'iguez et al.~\citep{sanchez2026understanding} applied LIME to BERT and MentalBERT to identify language patterns and their relation to predictions of four mental-health-related Reddit groups. However, the correctness and stability of such explanations remain essential methodological considerations~\citep{jain2019attention}. Explainability has also been integrated into transformer ensembles. Islam et al.~\citep{islam2026ensemble} utilized LIME on a validation-weighted ensemble of transformers for emotion and severity categorization of depression and assessed explanations by token deletion and stability analyses. Instead, explainability and calibrated confidence were included in ensemble weighting by Demirta\c{s} Alpsalaz et al.~\citep{alpsalaz2026explainability}. Based on classifier complementarity, attribution methods offer a complementary perspective on whether models that produce distinct correct predictions also rely on diverse linguistic evidence. Hence, explainability can be utilized diagnostically to interrogate the mechanisms behind successful fusion, and counterfactual disturbance provides a stronger test of whether observed attribution discrepancies are functionally related to ensemble decisions. Existing work supports domain adaptation, heterogeneous transformer ensembles, probability calibration, leakage-aware evaluation, and post-hoc explanation. But these aspects have been extensively investigated independently. In this work, we integrate these elements into a leakage-aware, calibration-conscious decision-level fusion framework and ask whether class-specific correctness complementarity measured on validation data is associated with fusion gain on an independent test set. Table~\ref{tab:related_compare} compares the proposed framework with representative prior works in transformer-based mental-health classification.}

\begin{table*}[!http]
\centering
\caption{Comparison of this work with representative prior studies on transformer-based mental-health text classification.}
\label{tab:related_compare}
\renewcommand{\arraystretch}{1.15}
\footnotesize
\begin{tabularx}{\textwidth}{@{}lcccccX@{}}
\toprule
Study & Multi-arch & Calibr.\ & Leakage & CI & XAI & Primary focus\\
 & ensemble   & reported & audit   & metric & link &                    \\
\midrule
Alsentzer et al.~\citep{alsentzer2019clinicalbert} & -- & -- & -- & -- & -- & Clinical-notes BERT embeddings\\
Ji et al.~\citep{ji2022mentalbert}                 & -- & -- & -- & -- & -- & MentalBERT/MentalRoBERTa release\\
Tavchioski et al.~\citep{tavchioski2023domain}      & partial & -- & -- & -- & -- & Reddit depression ensembles\\
Harrigian et al.~\citep{harrigian2021state}         & -- & -- & partial & -- & -- & Critique of data practices\\
Chancellor \& De Choudhury~\citep{chancellor2020methods} & -- & -- & -- & -- & -- & Methodological critique\\
Abdar et al.~\citep{abdar2021review}                & -- & \checkmark & -- & -- & -- & UQ in deep learning survey\\
\midrule
\textbf{This work}                                  & \checkmark & \checkmark & \checkmark & \checkmark & \checkmark & Leakage-aware calibrated decision-level fusion\\
\bottomrule
\end{tabularx}
\end{table*}

\begin{figure*}[!htbp]
\centering
\includegraphics[width=0.7\textwidth]{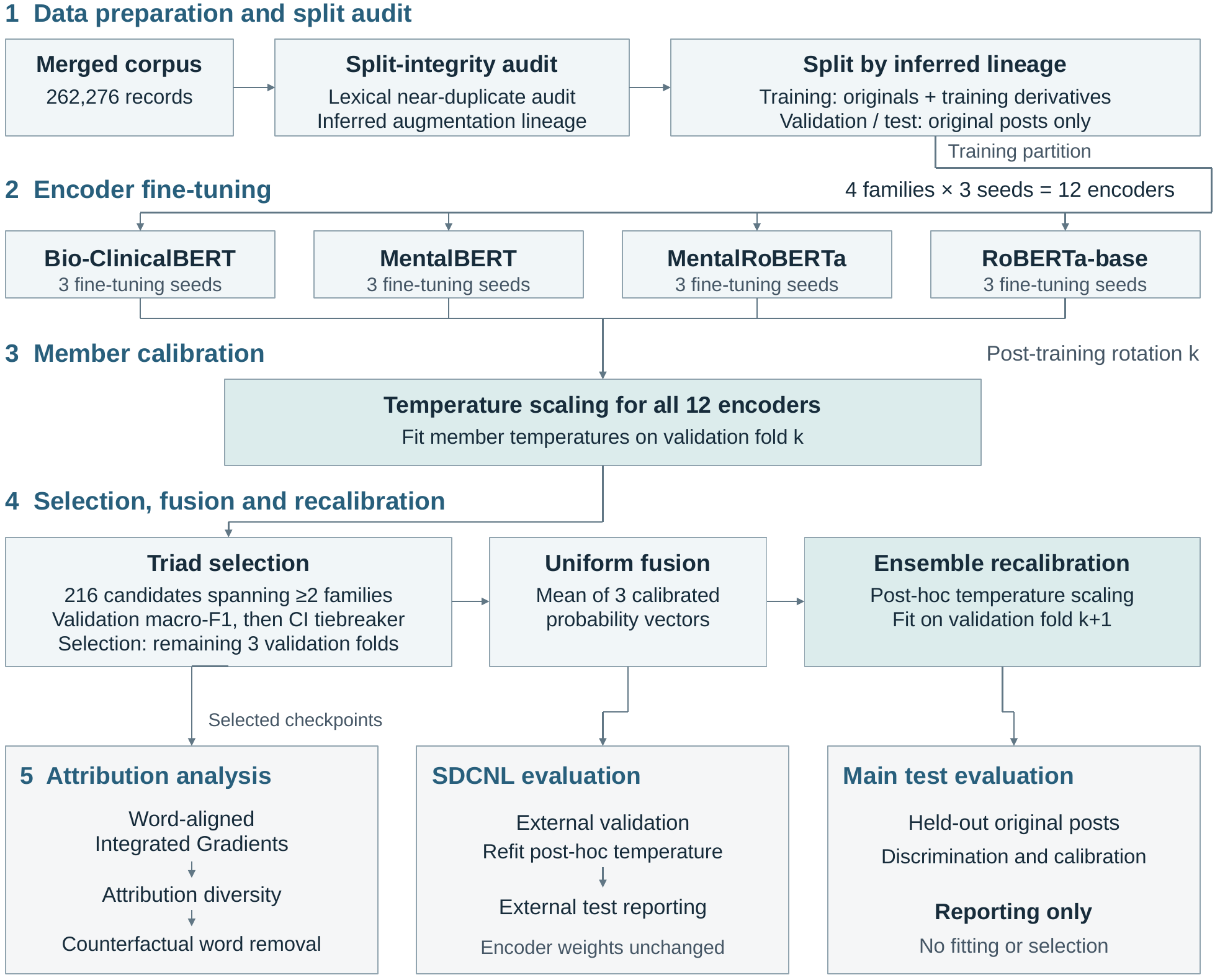}
\caption{Overview of the leakage-aware, cross-fitted calibration and complementarity-guided decision-level fusion framework.}
\label{fig:architecture}
\end{figure*}

\section{Theoretical Foundations}
\label{sec:theory}

This section formalizes the building blocks used in Section~\ref{sec:method}.

\subsection{Problem setup}
\label{subsec:setup}

Let $\mathcal{X}$ denote a space of text sequences and $\mathcal{Y}=\{1,\dots,K\}$ a finite label set; here $K=4$. A classifier is a map $f:\mathcal{X}\to\Delta^{K-1}$ where $\Delta^{K-1}$ is the probability simplex over $\mathcal{Y}$. Given $M$ trained classifiers $f_1,\dots,f_M$, a fusion rule produces a combined classifier $\tilde{f}:\mathcal{X}\to\Delta^{K-1}$. We target two objectives simultaneously: predictive accuracy, measured by macro-F1 and Matthews Correlation Coefficient (MCC~\citep{matthews1975mcc}); and confidence calibration, measured by Expected Calibration Error (ECE~\citep{naeini2015obtaining}) and Brier score.

\subsection{Temperature scaling}
\label{subsec:temperature}

Let $z=f_\theta^{\text{logit}}(x)\in\mathbb{R}^K$ denote the pre-softmax logits of a fine-tuned transformer. The softmax probability is $\sigma(z)_k = \frac{\exp(z_k)}{\sum_{j}\exp(z_j)}$. Temperature scaling~\citep{guo2017calibration} replaces $\sigma(z)$ with
\begin{equation}
  \sigma_T(z)_k \;=\; \frac{\exp(z_k/T)}{\sum_{j=1}^{K}\exp(z_j/T)},
  \qquad T\in\mathbb{R}_{>0},
  \label{eq:tempscale}
\end{equation}
where $T$ is chosen on a held-out validation set to minimize the negative log-likelihood:
\begin{equation}
  T^{\ast} \;=\; \arg\min_{T>0}\;
  -\sum_{(x,y)\in\mathcal{D}_{\text{val}}}
  \log\sigma_T\!\left(f_\theta^{\text{logit}}(x)\right)_y.
  \label{eq:tempfit}
\end{equation}
As a single-parameter, monotonic transform that leaves $\arg\max_k\sigma_T(z)_k$ invariant, temperature scaling adjusts confidence without changing hard predictions. We apply Eq.~\eqref{eq:tempfit} to each base model independently\revx{, on the validation fold reserved for member calibration (Section~\ref{subsec:protocol}). Because $\sigma_T(z)$ depends on $z$ only through differences of logits, the temperature can be fitted from stored softmax outputs by taking $z=\log\sigma(z)$ up to an additive constant.}

\subsection{Candidate Fusion Rules}
\label{subsec:fusion}

We evaluate both uniform probability averaging and validation-fitted per-class weighting as candidate decision-level fusion rules; the final selected system uses uniform averaging. We combine calibrated probabilities by a convex combination that is allowed to vary across classes:
\begin{equation}
  p_{\text{ens}}(y=c\mid x)
  \;=\;
  \frac{\sum_{m=1}^{M}\alpha_{m,c}\,\sigma_{T_m}\!\bigl(f_m^{\text{logit}}(x)\bigr)_c}
       {\sum_{c'=1}^{K}\sum_{m=1}^{M}\alpha_{m,c'}\,\sigma_{T_m}\!\bigl(f_m^{\text{logit}}(x)\bigr)_{c'}},
  \label{eq:ensemble}
\end{equation}
with per-class simplex constraints $\alpha_{m,c}\geq 0$ and $\sum_{m}\alpha_{m,c}=1$ for every $c$. We fit $\{\alpha_{m,c}\}$ on the validation set by coordinate-descent grid search on macro-F1. Uniform weights $\alpha_{m,c}=1/M$ recover the arithmetic-mean ensemble. Because the weights of each class are constrained to the simplex, the rule has $K(M-1)$ free parameters: $8$ for a three-member ensemble and $44$ for a twelve-member ensemble.
\FloatBarrier

\subsection{Complementarity Index (CI)}
\label{subsec:ci}

The CI is defined as a per-class measure of correctness complementarity between two trained classifiers. It is computed from validation-set predictions and used as a pre-fusion criterion for estimating the potential benefit of combining candidate encoders before probability aggregation.

For a class $c$ and two classifiers $A$ and $B$, CI is defined as
\begin{equation}
  \mathrm{CI}_c(A,B)
  =
  \frac{n^{A}_{c,\mathrm{only}} + n^{B}_{c,\mathrm{only}}}
       {n^{A\cup B}_{c}},
  \label{eq:ci}
\end{equation}
where $n^{A}_{c,\mathrm{only}}$ denotes the number of class-$c$ validation examples correctly classified only by $A$, $n^{B}_{c,\mathrm{only}}$ denotes the corresponding number for $B$, and $n^{A\cup B}_{c}$ denotes the number of class-$c$ validation examples correctly classified by at least one of the two classifiers. When $n^{A\cup B}_{c}=0$, we set $\mathrm{CI}_c(A,B)=0$.

Equivalently, let
\[
S_c^A=\{i:\hat{y}_A(x_i)=y_i=c\},
\qquad
S_c^B=\{i:\hat{y}_B(x_i)=y_i=c\}
\]
be the sets of class-$c$ validation examples correctly classified by $A$ and $B$, respectively. Then,
\begin{equation}
  \mathrm{CI}_c(A,B)
  =
  \frac{|S_c^A \triangle S_c^B|}
       {|S_c^A \cup S_c^B|}.
  \label{eq:ci_set}
\end{equation}
Equation~\eqref{eq:ci_set} is exactly the Jaccard distance between the two correctness sets, $\mathrm{CI}_c(A,B)=1-|S_c^A\cap S_c^B|/|S_c^A\cup S_c^B|$, with the convention $\mathrm{CI}_c=0$ when the union is empty. The contribution claimed here is therefore not the distance itself but its application: it is computed per class, on correctness sets rather than raw predictions, from validation data alone, and evaluated against fusion gain measured on an independent sample. A higher CI indicates that the two classifiers correctly handle more non-overlapping examples of class $c$, making their predictions more likely to provide useful complementary evidence under fusion. By construction, for a fixed union of correctly classified examples, CI increases with the non-overlap between the two correctness sets. Two properties limit what the index can express. It is computed from hard decisions and therefore carries no confidence information, although it anticipates the benefit of averaging probabilities; it also counts only examples of class $c$ that are classified correctly, so out-of-class false positives that also enter the class-$c$ F1 are outside its definition. Both are deliberate simplifications that keep the index computable from stored predictions alone, and both bound how much fusion gain it can explain.

For selecting ensembles with more than two members, candidate encoder sets are scored by averaging their pairwise CI values across all model pairs and classes:
\begin{equation}
  \mathrm{CI}(S)
  =
  \frac{1}{K\binom{|S|}{2}}
  \sum_{c=1}^{K}
  \sum_{\{A,B\}\subset S}
  \mathrm{CI}_c(A,B),
  \label{eq:ci_set_score}
\end{equation}
where $S$ denotes a candidate encoder set and $K$ is the number of classes. Candidate encoder sets are ranked by validation macro-F1, with $\mathrm{CI}(S)$ available as a tiebreaker among sets of comparable validation performance. Section~\ref{subsec:selection} reports how often that tiebreaker changes the selected ensemble under several tolerances; it is never decisive under the protocol used here, and the index is therefore evaluated as a predictor of fusion gain rather than as a selection mechanism. Ranking candidate sets by validation macro-F1 requires the candidate ensembles themselves to be formed, so the qualifier ``before aggregation'' applies to the computation of CI from member predictions, not to the ranking step. \revx{In Section~\ref{sec:results}, CI computed on validation predictions is evaluated as a predictor of the class-wise fusion gain $\Delta F_{1,c}$ realized on the test set, so that the predictor and the quantity predicted are measured on disjoint samples. The realized gain of a pair $\{A,B\}$ on class $c$ is defined as the class-$c$ F1 of the uniform-mean ensemble minus the larger of the two members' class-$c$ F1 values on the test set. This is a descriptive outcome variable, not a selection criterion: it is the quantity CI is asked to predict, and no model, weight, or temperature is chosen from it. Because the better member is identified per class on the same sample, it is the conservative choice, and Section~\ref{subsec:pairs} reports the alternative in which the reference is fixed in advance from validation data.}

\subsection{Post-hoc ensemble calibration}
\label{subsec:posthoc}

Averaging temperature-scaled members does not, in general, preserve calibration: the resulting aggregate distribution can remain systematically under- or over-confident. We therefore fit a second-stage temperature $T^{\star}$ directly to the aggregated prediction:
\begin{equation}
  p^{\star}(y\mid x)
  \;=\;
  \sigma_{T^{\star}}\!\bigl(\log p_{\text{ens}}(y\mid x)\bigr),
  \label{eq:posthocscale}
\end{equation}
with $T^{\star}$ minimizing the same negative log-likelihood in Eq.~\eqref{eq:tempfit} over \revx{the validation fold reserved for post-hoc calibration. A value $T^{\star}<1$ indicates that the ensemble of individually calibrated members is under-confident and is sharpened by the correction; $T^{\star}>1$ indicates over-confidence. Both regimes are observed in Section~\ref{sec:results}: $T^{\star}<1$ in domain and $T^{\star}>1$ under transfer to an external corpus.}

\subsection{Integrated Gradients on transformer inputs}
\label{subsec:ig}

Given an input embedding $\mathbf{e}(x)\in\mathbb{R}^{L\times d}$ where $L$ is the sequence length, and a baseline $\mathbf{e}_0$ (taken here as the zero embedding), the Integrated-Gradients attribution~\citep{sundararajan2017ig} of token $\ell$ for predicted class $\hat{y}$ is
\begin{equation}
  \text{IG}_\ell(x)
  \;=\;
  \bigl(\mathbf{e}(x)_\ell - \mathbf{e}_{0,\ell}\bigr)
  \cdot
  \int_{0}^{1}
   \frac{\partial F_{\hat{y}}\!\bigl(\mathbf{e}_0 + s(\mathbf{e}(x)-\mathbf{e}_0)\bigr)}{\partial \mathbf{e}_\ell}\;\mathrm{d}s,
  \label{eq:ig}
\end{equation}
where $F_{\hat{y}}$ is the scalar logit for class $\hat{y}$. We approximate Eq.~\eqref{eq:ig} by a Riemann sum with $20$ interpolation points (Sec.~\ref{sec:xai}) and reduce over the hidden dimension $d$ to obtain a per-token scalar $a_\ell(x)\in\mathbb{R}$. \revx{Because the ensemble members use different tokenizers (WordPiece for the BERT-family encoders, byte-pair encoding for the RoBERTa-family encoders), token-level attribution vectors of different members are not aligned position by position. Token attributions are therefore summed over the sub-word tokens whose character offsets fall inside each whitespace-delimited word, yielding a word-level attribution vector $\mathbf{a}^{w}(x)$ of common dimension for all members; inter-model comparisons in Section~\ref{sec:xai} are made on these word-level vectors, restricted to words covered by every member within the $256$-token window.}

\FloatBarrier

\section{Proposed Pipeline}
\label{sec:method}

\begin{revblock}
\subsection{Corpus construction}
\label{subsec:data}

The corpus was collected from five public collections, and its labels were mapped into four proxy categories: anxiety, depression, schizophrenia, and suicidal ideation. The sources were (i) the Kaggle \emph{Sentiment Analysis for Mental Health} corpus, which contains statements labeled as Normal, Depression, Suicidal, Anxiety, Bi-Polar, Stress, or Personality Disorder~\citep{ding2025tradeoffs}; (ii) the Dreaddit corpus of Reddit posts from five stress-related domains, including anxiety~\citep{turcan2019dreaddit}; (iii) the Kaggle \emph{Suicide and Depression Detection} corpus of $232{,}074$ Reddit posts labeled as suicide or non-suicide, collected from r/SuicideWatch and r/teenagers~\citep{nikhileswar2021suicide}; (iv) the Kaggle \emph{Depression: Reddit Dataset (Cleaned)}, which is a cleaned corpus of labeled Reddit posts; and (v) the Reddit Mental Health Dataset (RMHD), which contains posts from $28$ Reddit communities including $15$ specific mental-health support groups such as r/Anxiety, r/depression, r/schizophrenia, and r/SuicideWatch~\citep{low2020natural}. Labels were harmonized before merging: suicide and
suicidal ideation were mapped to suicidal ideation;
stress and anxiety to anxiety;
depression, depressive, bipolar, and
mental\_illness to depression; and
schizophrenia to schizophrenia. Records not mapped to one of these four target categories were excluded. The merged corpus kept only entries that mapped to the four target categories. All records are markers of a self-declared ailment, not a clinical diagnosis. The texts were made lowercase, contractions were extended, and non-alphabetic characters were eliminated. Exact duplicates were dropped. The merged file did not retain source IDs, author identifiers, or timestamps, which has ramifications as detailed in Section~\ref{subsec:threats}.

The merged file contains $262{,}276$ records, of which $149{,}195$ are original posts and $113{,}081$ are augmented derivatives produced from original anxiety, depression, and schizophrenia posts by lexical augmentation (synonym substitution and random word deletion, in the manner of EDA~\citep{wei2019eda}) to bring those classes to fixed target sizes (Table~\ref{tab:corpus}). Augmented records are marked by a \texttt{source} field, but the identity of the parent post was not recorded. The suicidal-ideation class was not augmented.

\begin{table}[!htbp]
\centering
\caption{Composition of the merged corpus by class and record type. Group size is the number of records (the original post plus its augmented derivatives) that share a parent post.}
\label{tab:corpus}
\footnotesize
\begin{adjustbox}{max width=\textwidth}
\begin{tabular}{@{}lrrrccc@{}}
\toprule
Class & Original & Augmented & Total & \multicolumn{3}{c}{Group size (mean / median / max)}\\
\midrule
Anxiety        & 5{,}210   & 34{,}790  & 40{,}000  & 7.7 & 8 & 20\\
Depression     & 16{,}209  & 43{,}791  & 60{,}000  & 3.7 & 3 & 16\\
Schizophrenia  & 500       & 34{,}500  & 35{,}000  & 70.0 & 68 & 136\\
Suicidal       & 127{,}276 & 0         & 127{,}276 & 1.0 & 1 & 1\\
\midrule
Total          & 149{,}195 & 113{,}081 & 262{,}276 & & & \\
\bottomrule
\end{tabular}
\end{adjustbox}
\end{table}

\subsection{Split-integrity audit}
\label{subsec:audit}

\paragraph{Lexical near-duplicate audit}
A stratified random $80/10/10$ split of the merged file (seed $42$) yields $209{,}820$ training, $26{,}228$ validation, and $26{,}228$ test records. This split was audited with a standard MinHash--LSH pipeline~\citep{broder1997minhash,leskovec2014massive}: each record was shingled into $5$-token windows, encoded into a $128$-permutation MinHash signature, and indexed with a Jaccard threshold of $0.8$; a validation or test record was flagged if any training record exceeded the threshold. The audit flagged $7{,}622/26{,}228$ ($29.06\%$) test records and $7{,}623/26{,}228$ ($29.06\%$) validation records.

\paragraph{Lineage audit}
Lexical near-duplicate detection identifies derivatives that remain close to their parent in $5$-gram space; it does not identify derivatives whose wording has drifted further, nor an original test post whose derivatives lie in the training set. To quantify both, the parent of every augmented record was recovered as the original post of the same class with the highest TF--IDF cosine similarity. Parent identity was inferred, not recorded: for each augmented record, the most similar original of the same class was taken as its parent. The first percentile of parent similarity was $0.87$ (anxiety), $0.89$ (depression), and $0.83$ (schizophrenia), the median exceeded $0.98$ in every class, and $9$ of $113{,}081$ records fell below $0.5$. These figures show that almost every derivative has a highly similar original of the same class, not that the specific original recovered is the true parent; where several originals are near-identical the assignment may be wrong, and the split is lineage-disjoint with respect to the inferred lineage. With lineage known, a test record was defined as lineage-linked if its lineage group (the parent post together with all of its derivatives) has at least one member in the training set. A lineage-linked record is therefore either an augmented derivative of a training post or an original post whose derivatives lie in the training partition; both forms create the same dependence between the partitions. Table~\ref{tab:leakage} and Fig.~\ref{fig:leakage} report the result for the random split: $13{,}164/26{,}228$ ($50.2\%$) test records are lineage-linked, comprising every schizophrenia record, $99.5\%$ of anxiety records, and $94.7\%$ of depression records; the suicidal-ideation class, which was never augmented, is unaffected. Of the $13{,}164$ lineage-linked test records, $11{,}219$ are augmented derivatives of training posts and $1{,}945$ are original posts whose derivatives lie in training. Of the $18{,}606$ test records that the lexical audit retains, $5{,}588$ ($30.0\%$) remain lineage-linked: $5{,}282$ augmented records and $306$ originals. The lexical audit therefore removes the near-verbatim contamination layer and leaves the paraphrase layer in place.

\begin{table}[!htbp]
\centering
\caption{Two-layer audit of the stratified random test split ($N=26{,}228$). ``Lexical'' is the MinHash--LSH audit (Jaccard $\geq0.8$ against the training set); ``Lineage-linked'' records share an augmentation lineage group with at least one training record, either as derivatives of a training post or as originals whose derivatives lie in training.}
\label{tab:leakage}
\footnotesize
\setlength{\tabcolsep}{4pt}
\begin{adjustbox}{max width=\textwidth}
\begin{tabular}{@{}lrrrr@{}}
\toprule
Class & Test records & Lexically flagged & lineage-linked & Retained by lexical audit, lineage-linked\\
\midrule
Anxiety       & 4{,}000  & \multirow{4}{*}{7{,}622 (29.1\%)} & 3{,}980 (99.5\%) & \multirow{4}{*}{5{,}588 of 18{,}606 (30\%)}\\
Depression    & 6{,}000  &  & 5{,}684 (94.7\%) & \\
Schizophrenia & 3{,}500  &  & 3{,}500 (100\%)  & \\
Suicidal      & 12{,}728 &  & 0 (0\%)          & \\
\midrule
Total         & 26{,}228 & 7{,}622 (29.1\%) & 13{,}164 (50.2\%) & 5{,}588 (30\%)\\
\bottomrule
\end{tabular}
\end{adjustbox}
\end{table}

\begin{figure}[!htbp]
\centering
\includegraphics[width=0.8\columnwidth]{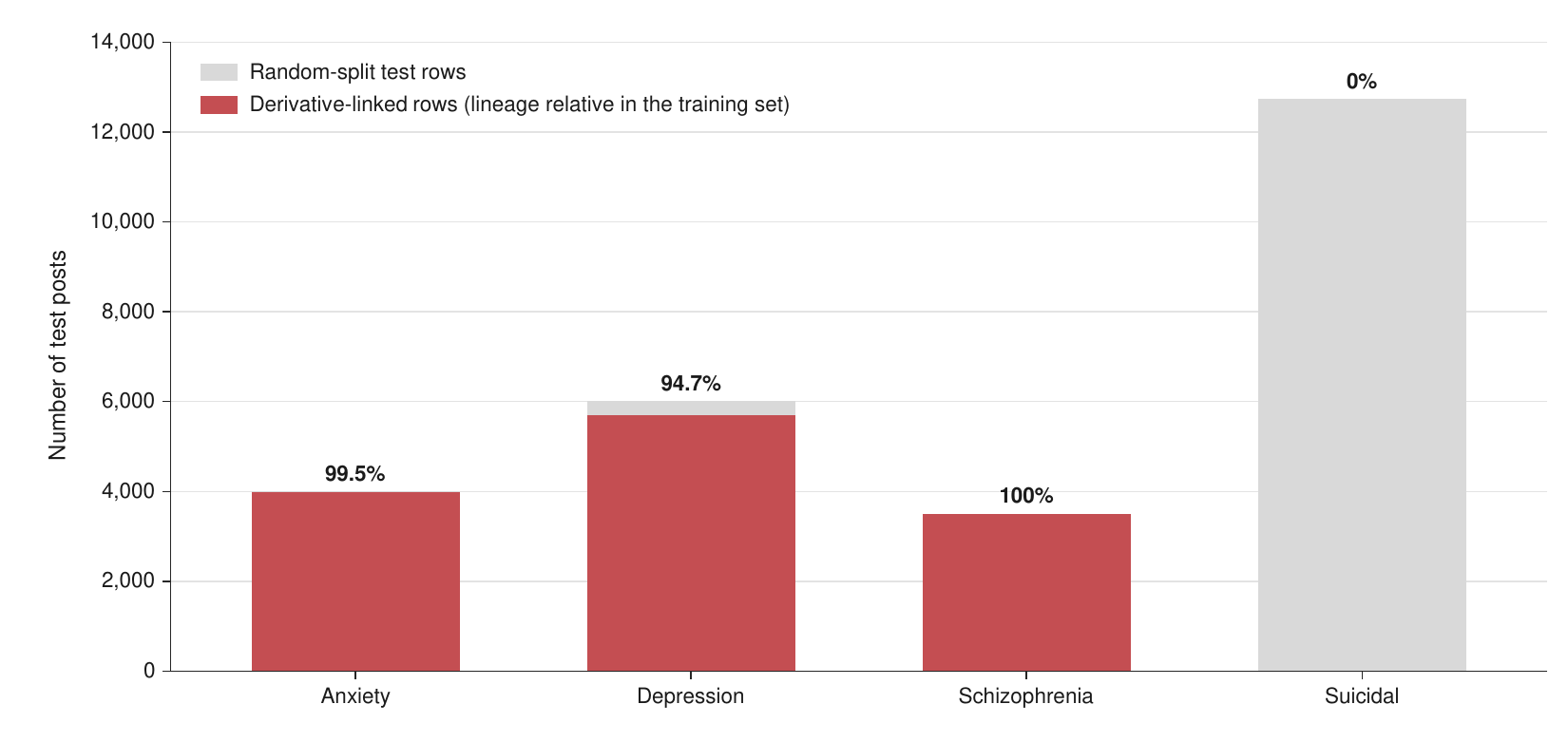}
\caption{lineage-linked records in the stratified random test split, by class. The three augmented classes are almost entirely linked to training posts through augmentation lineage; the suicidal-ideation class, which was not augmented, is unaffected.}
\label{fig:leakage}
\end{figure}

\paragraph{Lineage-disjoint split}
Because the contamination is structural, it was removed by construction rather than by filtering. The $149{,}195$ original posts were split $70/15/15$ with class stratification (seed $42$); every augmented record was then assigned to the split of its parent, and augmented records whose parent fell in the validation or test partition were discarded. Validation and test sets therefore contain original posts only; the training set contains original posts together with the derivatives of training posts only, and no lineage group spans two partitions (verified programmatically). A $70/15/15$ ratio was preferred to $80/10/10$ so that the smallest class retains $75$ test records rather than $50$. Table~\ref{tab:split} gives the resulting partition sizes.

\begin{table}[!htbp]
\centering
\caption{Lineage-disjoint split. Training contains original posts and their derivatives; validation and test contain original posts only.}
\label{tab:split}
\footnotesize
\setlength{\tabcolsep}{4pt}
\begin{adjustbox}{max width=\textwidth}
\begin{tabular}{@{}lrrrrr@{}}
\toprule
Partition & Anxiety & Depression & Schizophrenia & Suicidal & Total\\
\midrule
Train (original)   & 3{,}647 & 11{,}346 & 350 & 89{,}093 & 104{,}436\\
Train (augmented)  & 24{,}195 & 30{,}794 & 24{,}333 & 0 & 79{,}322\\
Train (total)      & 27{,}842 & 42{,}140 & 24{,}683 & 89{,}093 & 183{,}758\\
Validation         & 782 & 2{,}431 & 75 & 19{,}091 & 22{,}379\\
Test               & 781 & 2{,}432 & 75 & 19{,}092 & 22{,}380\\
\bottomrule
\end{tabular}
\end{adjustbox}
\end{table}

\paragraph{Residual lexical and semantic audits}
The same MinHash--LSH audit was then applied to the lineage-disjoint split. It flags $78/22{,}379$ ($0.35\%$) validation records and $83/22{,}380$ ($0.37\%$) test records against the training set, and $22$ test records against the validation set; inspection shows that these are re-posted or templated texts among original posts. Because a lexical audit cannot detect re-posts whose wording has been edited beyond the $5$-gram threshold, a semantic audit was added. Every validation and test post and every original training post was embedded with a sentence encoder (all-MiniLM-L6-v2, mean pooling, $\ell_2$-normalized), and the maximum cosine similarity to any original training post was recorded. At a threshold of $0.90$ the semantic audit flags $175$ validation ($0.78\%$) and $174$ test ($0.78\%$) posts, of which $73$ and $80$ respectively are also flagged lexically; $102$ validation and $94$ test posts are flagged semantically only ($109$ and $112$ test posts exceed $0.95$; $58$ and $65$ exceed $0.98$). The highest-scoring cases are re-posts that share a long verbatim prefix with a training post but diverge later. The median test post has a maximum similarity of $0.74$ to the training set and the $99$th percentile is $0.87$, so the flagged set is a thin tail rather than a shoulder of the distribution. The test retention mask therefore excludes $105$ records ($83$ matching training and $22$ matching validation). Of the $174$ test records flagged semantically, $80$ are already excluded by that mask, so the union of the two audits removes $199$ records and the union-strict test subset contains $22{,}181$ posts. Results are reported on the full test set; on the union-strict subset the headline macro-F1 is $0.8440$ versus $0.8436$ and the best single encoder $0.8233$ versus $0.8225$, so neither residual layer affects any conclusion.

\subsection{Encoders and fine-tuning}
\label{subsec:training}

Four encoder families were fine-tuned with three random seeds each ($42$, $43$, $44$), giving twelve models: Bio-ClinicalBERT~\citep{alsentzer2019clinicalbert} (pretrained on clinical notes), MentalBERT and MentalRoBERTa~\citep{ji2022mentalbert} (pretrained on mental-health Reddit text), and RoBERTa-base~\citep{liu2019roberta} (general web text). All runs share one recipe: a sequence-classification head on the pretrained backbone, the lower six of the twelve encoder layers frozen and the upper six fine-tuned, hidden dropout $0.3$, maximum sequence length $256$, batch size $16$, the Adam optimizer with learning rate $2\times10^{-5}$ and a reduce-on-plateau schedule (factor $0.2$, patience $1$ epoch), class-balanced sample weights computed on the training partition, at most $10$ epochs, and early stopping on validation loss with patience $3$ and restoration of the best weights. Implementation used TensorFlow $2.15.1$ and Transformers $4.40.2$. Each run took between $4$ and $12$\,h on an NVIDIA A100-80GB GPU. Validation loss is used for early stopping and checkpoint selection, and the validation set is therefore not independent of the trained weights; every subsequent use of validation data is confined to the cross-fitted protocol described next.

\subsection{Evaluation protocol}
\label{subsec:protocol}

\paragraph{Cross-fitted post-training decisions}
Four post-training quantities depend on validation data: the member temperatures $T_m$, the ensemble selection (validation macro-F1 among triads spanning at least two encoder families, with $\mathrm{CI}(S)$ as a tiebreaker among candidates of equal validation macro-F1), the optional per-class weights $\{\alpha_{m,c}\}$, and the post-hoc temperature $T^{\star}$. To prevent any of these from being tuned on the same records, the validation set was partitioned into $K=5$ folds. In rotation $k$, the member temperatures were fitted on fold $k$, the ensemble was selected on the three folds other than $k$ and $k{+}1$, and the post-hoc temperature was fitted on fold $k{+}1$ (indices modulo $5$). Uniform averaging is the pre-specified fusion rule (Section~\ref{subsec:fusion}); the learned per-class weights are evaluated as an alternative within the same rotations, fitted on the selection folds and scored on the held-out fold $k{+}1$, so that the choice between the two rules rests on validation data alone (Section~\ref{subsec:nway}). Each rotation yields a complete system that is scored on the test set; the mean and spread across rotations constitute the primary, post-selection estimate of the procedure's performance. The system selected on the full validation set under the same rule is reported as the deployed instance. The test set is used for reporting only: no temperature, weight, threshold, or selection decision in the study was informed by test-set results. 

\paragraph{Metrics and uncertainty}
Macro-F1, accuracy, balanced accuracy, MCC~\citep{matthews1975mcc}, ECE with $15$ equal-width bins~\citep{naeini2015obtaining}, and the multiclass Brier score are reported. Confidence intervals for macro-F1 and MCC are $95\%$ percentile intervals from $B=1000$ non-parametric bootstrap resamples of the test set~\citep{efron1986bootstrap}; per-class F1 intervals are computed in the same way. ECE and Brier score are reported as point estimates on the full test set together with $95\%$ bootstrap intervals, and paired bootstrap intervals are given for the differences in ECE and Brier score between compared systems; the sensitivity of ECE to the estimator is examined with $10$, $15$, $20$, and $30$ equal-width bins, $15$ equal-mass bins, a class-wise variant~\citep{nixon2019measuring}, and a binning-free kernel estimator~\citep{popordanoska2022consistent}.

\paragraph{Significance testing}
Head-to-head comparisons use McNemar's test with continuity correction~\citep{mcnemar1947} at the level of individual predictions, a paired bootstrap of the macro-F1 difference ($B=1000$), and a permutation test in which the predictions of the two systems are exchanged per record ($2{,}000$ permutations). In every McNemar test, $b$ counts records the reference system classifies correctly, and the compared system does not, and $c$ counts the reverse; $c>b$ therefore favors the compared system. All ensemble--versus--member McNemar tests performed in the study ($66$ pairwise ensembles and the headline comparison, $67$ tests) form one family, and the headline $p$-value is reported after Holm correction over that family~\citep{holm1979simple}.

\paragraph{Evaluating CI as a predictor of fusion gain}
CI is computed on the full validation set and the realized gain on the test set for all $\binom{12}{2}=66$ pairs and $4$ classes ($264$ pair--class observations). Because observations sharing a pair or a class are not independent, the pooled correlation is accompanied by cluster bootstraps over pairs and over models ($2{,}000$ resamples), within-class correlations, a linear mixed-effects model, and leave-one-encoder-out prediction of gain from CI. Because the $66$ pairs are formed from only $12$ encoders, each of which appears in $11$ pairs, pair-level resampling alone understates the dependence. Within-class correlations are also reported with an encoder-level bootstrap, in which encoders are resampled with replacement, and each pair is weighted by the product of its two members' multiplicities, and with a leave-one-encoder-out recomputation. The mixed-effects model uses a symmetric multi-membership random intercept. A single random effect per encoder enters every pair containing it, together with a random intercept per pair. Per-class slopes are estimated with the encoder effect alone, because with one observation per pair and class, a pair intercept would identify the slope from within-pair variation across classes. Two definitions are used in the paper. In the pairwise and head-to-head comparisons of Tables~\ref{tab:pairs_top} and~\ref{tab:head2head}, where a system is compared with a baseline a practitioner could have deployed, the reference is the member with the higher validation macro-F1, so that no baseline is chosen on the test set. In the CI analysis, where the gain is the outcome being predicted rather than a claimed improvement, the reference is the better of the two members on the given class (Section~\ref{subsec:ci}); results under the validation-selected reference are reported alongside. The CI--gain association and the tolerance comparisons of Table~\ref{tab:ci_ablation} were not pre-specified and are reported as exploratory; no test-set result informed any fitting or selection decision.

\paragraph{Lexical baseline and artifact controls}
A TF--IDF (unigram and bigram) logistic-regression classifier with class-balanced weights was trained on the lineage-disjoint training set to measure the lexically separable part of the task. Reliance on platform artifacts was tested by re-scoring the test set with Reddit meta-vocabulary masked (e.g., reddit, subreddit, mods, tldr, edit, throwaway, tw) and, additionally, with explicit condition names and their derivatives masked (e.g., anxiety, depressed, schizophrenia, antipsychotics, suicidal); performance is also reported on the natural subset of test posts that contain no condition name.

\paragraph{External corpus}
Zero-shot transfer was evaluated on SDCNL~\citep{haque2021sdcnl}, an independently collected corpus of $1{,}895$ posts that retains author identifiers. Posts were cleaned with the same procedure as the main corpus; $4$ posts whose body was deleted or removed retained fewer than five tokens after cleaning and were discarded, leaving $1{,}891$. Of these, the $83$ posts ($4.4\%$) that the MinHash--LSH audit flagged as near-duplicates of any main-corpus record were removed, leaving $1{,}808$, which were divided into author-disjoint halves of $892$ and $916$.

\paragraph{Attribution analysis}
Integrated Gradients (Eq.~\eqref{eq:ig}; $20$ interpolation steps, zero baseline) were computed for the three deployed checkpoints of the headline ensemble on $1{,}575$ test posts: all $75$ schizophrenia posts and $500$ posts sampled from each other class (seed $0$). Agreement between the attribution run's predictions and the stored test predictions was verified for every checkpoint ($\geq 99.9\%$ identical arg-max decisions). Word-level alignment follows Section~\ref{subsec:ig}. The counterfactual protocol is described in Section~\ref{sec:xai}.
\end{revblock}

\FloatBarrier

\section{Results}
\label{sec:results}

\begin{revblock}
\subsection{Individual encoders}
\label{subsec:individual}

Table~\ref{tab:individual} reports the twelve fine-tuned encoders on the lineage-disjoint test set ($N=22{,}380$), after member-wise temperature scaling, and Fig.~\ref{fig:indiv_per_class} shows per-class F1 with bootstrap intervals. Three observations follow. First, seed-averaged macro-F1 orders the families as MentalRoBERTa ($0.832\pm0.009$), RoBERTa-base ($0.822\pm0.020$), MentalBERT ($0.819\pm0.007$), and Bio-ClinicalBERT ($0.803\pm0.005$), but the $95\%$ intervals of individual runs are approximately $\pm0.02$ wide and overlap across families; the difference between the mental-health-specialized families and Bio-ClinicalBERT ($0.02$--$0.03$) is the only family contrast that is consistent across all seeds. Second, seed-to-seed variation is material: RoBERTa-base spans $0.799$--$0.835$ across seeds, a range larger than most between-family differences, which is why three seeds per family are used throughout. Third, per-class performance is highly uneven. Suicidal ideation, the largest class, is classified with F1 $0.963$--$0.968$ by every model; anxiety with $0.84$--$0.89$; depression, the class with the greatest lexical overlap with suicidal ideation, with $0.71$--$0.75$; and schizophrenia, with $75$ test posts, with $0.62$--$0.77$ and intervals roughly $\pm0.08$ wide. Member temperatures $T_m$ lie between $1.32$ and $1.52$, i.e.\ every encoder is over-confident before scaling, and post-scaling ECE is below $0.011$ for every model.

\begin{table*}[!htbp]
\centering
\caption{Individual-model performance on the lineage-disjoint test set ($N=22{,}380$) after member-wise temperature scaling. Macro-F1 and MCC are given with $95\%$ bootstrap percentile intervals ($B=1000$); ECE (15 bins), Brier score, accuracy, and balanced accuracy are point estimates on the full test set. Best value per column in \textbf{bold}.}
\label{tab:individual}
\renewcommand{\arraystretch}{1.15}
\footnotesize
\begin{adjustbox}{max width=\textwidth}
\begin{tabular}{@{}lccccccc@{}}
\toprule
Model & $T_m$ & Macro-F1 & MCC & ECE & Brier & Accuracy & Bal.\ acc.\\
\midrule
Bio-ClinicalBERT\,s42 & 1.47 & 0.8064 [0.7852, 0.8265] & 0.7382 [0.7262, 0.7489] & 0.0056 & 0.0973 & 0.9298 & 0.8494\\
Bio-ClinicalBERT\,s43 & 1.49 & 0.7968 [0.7755, 0.8177] & 0.7411 [0.7295, 0.7529] & 0.0039 & 0.0952 & 0.9312 & 0.8330\\
Bio-ClinicalBERT\,s44 & 1.52 & 0.8056 [0.7810, 0.8261] & 0.7400 [0.7280, 0.7514] & 0.0082 & 0.0956 & 0.9311 & 0.8408\\
MentalBERT\,s42       & 1.43 & 0.8107 [0.7884, 0.8309] & 0.7538 [0.7427, 0.7648] & 0.0109 & 0.0965 & 0.9306 & 0.8511\\
MentalBERT\,s43       & 1.33 & 0.8247 [0.8027, 0.8444] & 0.7585 [0.7474, 0.7696] & 0.0057 & 0.0911 & 0.9366 & 0.8372\\
MentalBERT\,s44       & 1.39 & 0.8219 [0.7987, 0.8423] & 0.7649 [0.7532, 0.7762] & 0.0043 & 0.0897 & 0.9367 & \textbf{0.8537}\\
MentalRoBERTa\,s42    & 1.32 & \textbf{0.8379 [0.8161, 0.8556]} & \textbf{0.7691 [0.7580, 0.7802]} & \textbf{0.0038} & \textbf{0.0861} & \textbf{0.9387} & 0.8426\\
MentalRoBERTa\,s43    & 1.39 & 0.8368 [0.8154, 0.8553] & 0.7641 [0.7527, 0.7753] & 0.0071 & 0.0897 & 0.9349 & 0.8472\\
MentalRoBERTa\,s44    & 1.37 & 0.8225 [0.8012, 0.8429] & 0.7648 [0.7534, 0.7754] & 0.0066 & 0.0889 & 0.9362 & 0.8474\\
RoBERTa-base\,s42     & 1.45 & 0.7994 [0.7761, 0.8204] & 0.7558 [0.7450, 0.7675] & 0.0107 & 0.0927 & 0.9333 & 0.8496\\
RoBERTa-base\,s43     & 1.42 & 0.8352 [0.8131, 0.8557] & 0.7684 [0.7574, 0.7797] & 0.0097 & 0.0883 & 0.9377 & 0.8450\\
RoBERTa-base\,s44     & 1.39 & 0.8307 [0.8070, 0.8509] & 0.7568 [0.7454, 0.7673] & 0.0095 & 0.0930 & 0.9322 & 0.8476\\
\bottomrule
\end{tabular}
\end{adjustbox}
\end{table*}

\begin{figure*}[!htbp]
\centering
\includegraphics[width=\textwidth]{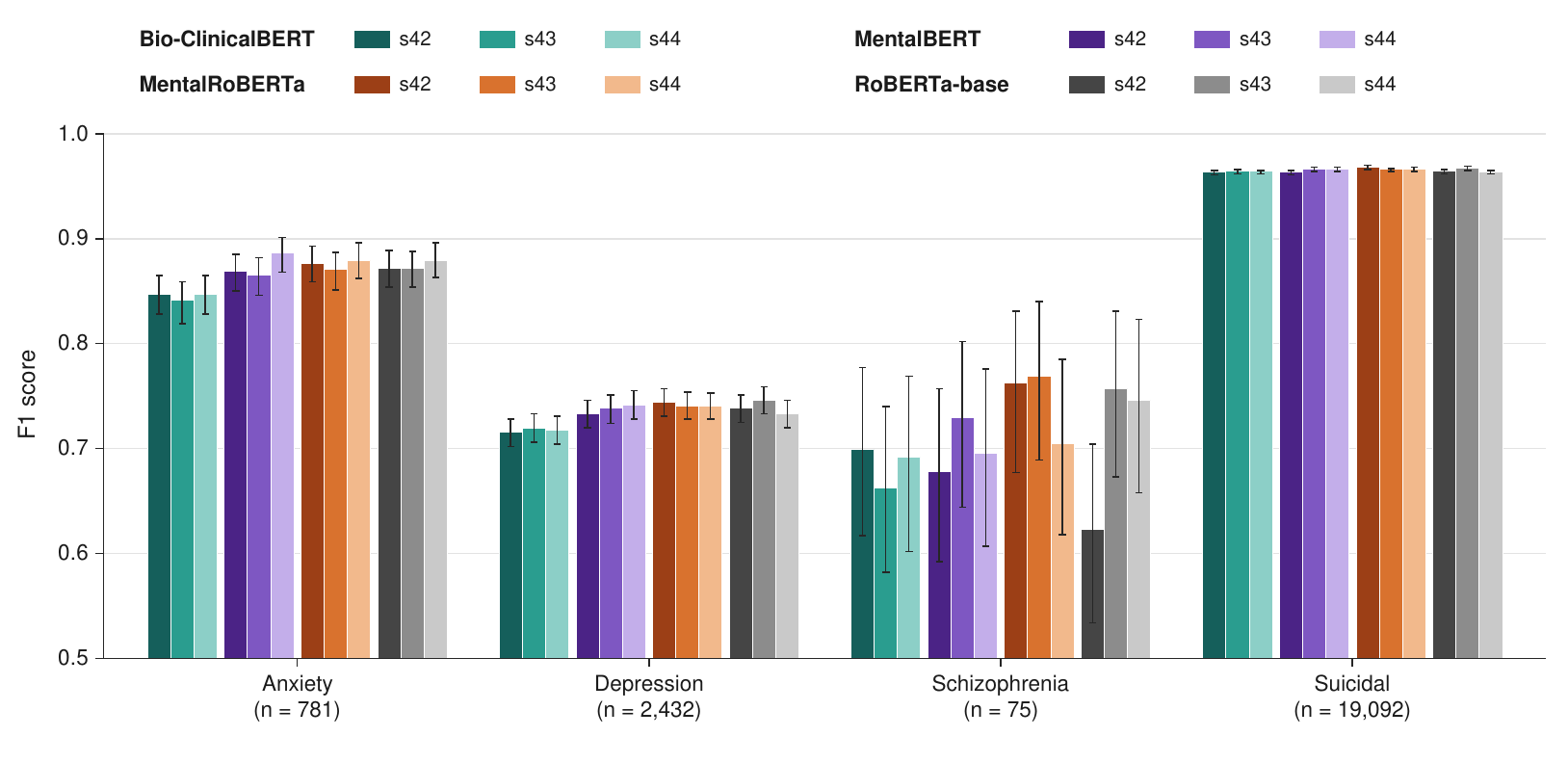}
\caption{Per-class F1 of the twelve fine-tuned encoders on the lineage-disjoint test set; error bars are $95\%$ bootstrap intervals. Depression is the hardest of the three large classes; schizophrenia has $75$ test posts and correspondingly wide intervals.}
\label{fig:indiv_per_class}
\end{figure*}

\subsection{Pairwise fusion and the Complementarity Index}
\label{subsec:pairs}

All $66$ pairs were fused by uniform averaging of calibrated probabilities. Table~\ref{tab:pairs_top} lists the ten strongest pairs by test macro-F1 together with the gain over a reference member; to avoid selecting the reference on the test set, the reference member of each pair is the one with the higher validation macro-F1. Fusion improves macro-F1 over the validation-selected member in $43$ of $66$ pairs; the paired-bootstrap interval of the gain excludes zero in the positive direction for $15$ pairs and in the negative direction for $3$. McNemar's test, which operates on per-record accuracy, rejects the null for $48$ of $66$ pairs; the discrepancy with the macro-F1 intervals arises because accuracy is dominated by the suicidal-ideation class ($85\%$ of records), whereas macro-F1 is governed by the three smaller classes. Both quantities are therefore reported throughout, and macro-F1 intervals are treated as the primary criterion.

\begin{table*}[!htbp]
\centering
\scriptsize
\caption{Ten strongest pairwise ensembles (uniform mean of calibrated probabilities) on the lineage-disjoint test set. The reference member is the member with the higher validation macro-F1; $\Delta$ is the difference between the ensemble's test macro-F1 and that of the reference member in Table~\ref{tab:individual}, with a $95\%$ paired-bootstrap interval ($B=2000$ resamples of the test set); $\overline{\mathrm{CI}}$ is the mean validation CI across classes; McNemar $p$ compares the ensemble with the reference member at the record level.}
\label{tab:pairs_top}
\renewcommand{\arraystretch}{1.15}
\setlength{\tabcolsep}{3pt}
\begin{adjustbox}{max width=\textwidth}
\begin{tabular}{@{}llcccc@{}}
\toprule
Pair & Reference (by validation) & macro-F1 & $\Delta$ vs.\ reference & $\overline{\mathrm{CI}}$ & McNemar $p$\\
\midrule
MentalBERT\,s43 + RoBERTa-base\,s44        & MentalBERT\,s43    & 0.8447 & $+0.0200$ [$+0.0076$, $+0.0329$] & 0.112 & $6.5\times10^{-3}$\\
MentalRoBERTa\,s43 + RoBERTa-base\,s43     & MentalRoBERTa\,s43 & 0.8443 & $+0.0075$ [$-0.0047$, $+0.0200$] & 0.082 & $1.5\times10^{-5}$\\
MentalRoBERTa\,s43 + RoBERTa-base\,s44     & MentalRoBERTa\,s43 & 0.8436 & $+0.0069$ [$-0.0016$, $+0.0166$] & 0.093 & $0.17$\\
MentalRoBERTa\,s42 + RoBERTa-base\,s44     & MentalRoBERTa\,s42 & 0.8422 & $+0.0043$ [$-0.0084$, $+0.0174$] & 0.090 & $0.66$\\
MentalBERT\,s42 + MentalRoBERTa\,s43       & MentalRoBERTa\,s43 & 0.8410 & $+0.0042$ [$-0.0080$, $+0.0185$] & 0.074 & $7.0\times10^{-3}$\\
MentalBERT\,s43 + MentalRoBERTa\,s43       & MentalRoBERTa\,s43 & 0.8404 & $+0.0036$ [$-0.0075$, $+0.0154$] & 0.096 & $2.1\times10^{-9}$\\
Bio-ClinicalBERT\,s43 + MentalRoBERTa\,s43 & MentalRoBERTa\,s43 & 0.8403 & $+0.0035$ [$-0.0105$, $+0.0192$] & 0.109 & $2.4\times10^{-5}$\\
RoBERTa-base\,s43 + RoBERTa-base\,s44      & RoBERTa-base\,s43  & 0.8401 & $+0.0049$ [$-0.0040$, $+0.0143$] & 0.085 & $2.2\times10^{-2}$\\
Bio-ClinicalBERT\,s42 + MentalRoBERTa\,s42 & MentalRoBERTa\,s42 & 0.8389 & $+0.0009$ [$-0.0142$, $+0.0161$] & 0.103 & $1.2\times10^{-2}$\\
MentalRoBERTa\,s42 + MentalRoBERTa\,s43    & MentalRoBERTa\,s42 & 0.8383 & $+0.0004$ [$-0.0097$, $+0.0091$] & 0.059 & $0.82$\\
\bottomrule
\end{tabular}
\end{adjustbox}
\end{table*}

Fig.~\ref{fig:ci_vs_gain} relates the CI, computed on validation predictions, to the fusion gain realized on the test set for the $264$ pair--class observations. CI values on this corpus lie between $0.02$ and $0.21$: the encoders make substantially complementary errors. The pooled Pearson correlation across all observations is $r=0.05$ (cluster-bootstrap interval over pairs $[-0.03,0.14]$; over models, with pairs weighted by the product of the sampled model multiplicities, $[-0.13,0.23]$), and this pooled figure is not informative because the four classes occupy different regions of both axes. The within-class picture is heterogeneous (Fig.~\ref{fig:ci_vs_gain}b): CI is positively associated with gain for depression ($r=0.50$) and suicidal ideation ($r=0.68$), the two most confusable classes, is uncorrelated with gain for anxiety ($r=-0.20$, interval including zero), and carries no usable signal for schizophrenia ($r=0.16$), whose per-pair gains have a standard deviation of $0.022$---five times that of any other class---because they are computed from $75$ posts. Excluding schizophrenia, the pooled correlation is $0.61$. A linear mixed-effects model with class as a fixed effect, a symmetric multi-membership random intercept per encoder, and a random intercept per pair estimates a CI slope of $0.062$ ($95\%$ interval $[-0.001,0.125]$, $p=0.053$) over all classes and $0.048$ ($[0.022,0.075]$, $p=0.0004$) over the three adequately sampled classes; the pooled slope is therefore not significant once schizophrenia is included, and a validation-measured accuracy gap between the two members adds no explanatory power. Under encoder-level resampling the within-class associations for depression and suicidal ideation remain positive and exclude zero ($r=0.50$, $[0.19,0.79]$; $r=0.68$, $[0.51,0.85]$), as do their per-class slopes with the encoder random effect ($0.076$ $[0.046,0.106]$ and $0.150$ $[0.112,0.188]$, both $p<0.001$), whereas the intervals for anxiety ($r=-0.20$, $[-0.50,0.27]$; slope $-0.029$ $[-0.113,0.055]$) and schizophrenia ($r=0.16$, $[-0.32,0.55]$; slope $0.073$ $[-0.127,0.273]$) contain zero. Dropping any single encoder and recomputing leaves every within-class correlation on the same side of zero (depression $0.40$--$0.55$, suicidal ideation $0.65$--$0.73$). Defining the gain instead against the member with the higher validation macro-F1 raises the pooled correlation to $r=0.15$ (cluster-bootstrap interval over pairs $[0.07,0.23]$) and the all-class slope to $0.121$ $([0.021,0.221]$, $p=0.018)$, while the slope over the three sampled classes is essentially unchanged at $0.041$ $([0.006,0.076]$, $p=0.023)$; the class-conditional pattern is the same under both definitions. Leave-one-encoder-out prediction, in which the regression is fitted on the pairs that exclude a given encoder and applied to the pairs containing it, gives pooled out-of-sample Spearman gives pooled out-of-sample Spearman $\rho=0.60$ and $R^{2}=0.27$; by class it is $\rho=0.66$ ($R^{2}=0.42$) for suicidal ideation and $\rho=0.38$ ($R^{2}=0.17$) for depression, and negative for anxiety and schizophrenia. CI therefore transfers to pairs containing an encoder it was not fitted on, but only for the two classes where the association exists. CI is therefore a class-conditional diagnostic rather than a global predictor: it indicates which classes will benefit from fusion, not by how much and not which pair to choose, and it is uninformative for a class whose test support is too small to measure gain reliably.

\begin{figure}[!htbp]
\centering
\includegraphics[width=\columnwidth]{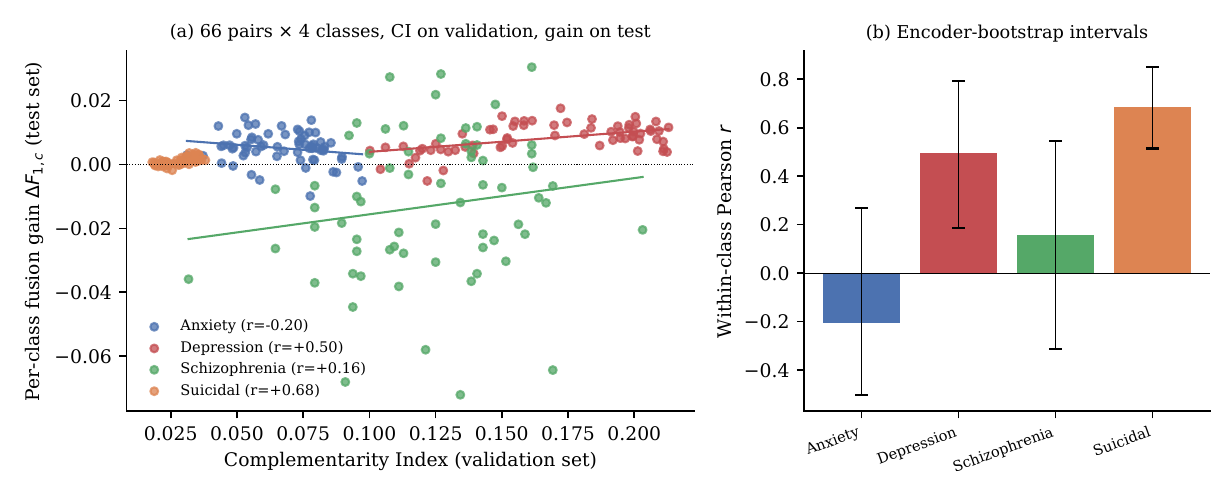}
\caption{Complementarity Index versus realized fusion gain. (a) CI computed on the validation set against the class-wise F1 gain of the uniform-mean pair over the better member's class-wise F1 on the test set (Section~\ref{subsec:ci}), for all $66$ pairs and $4$ classes, with per-class least-squares fits. (b) Within-class Pearson correlation with $95\%$ encoder-level bootstrap intervals (encoders resampled with replacement, each pair weighted by its member multiplicities). The association is positive for depression and suicidal ideation, absent for anxiety, and unmeasurable for the $75$-record schizophrenia class.}
\label{fig:ci_vs_gain}
\end{figure}

\subsection{Ensemble selection under the cross-fitted protocol}
\label{subsec:selection}

Three-member ensembles were selected from the $216$ candidate triads that span at least two encoder families, ranked by validation macro-F1 with $\mathrm{CI}(S)$ as a tiebreaker. Table~\ref{tab:crossfit} reports the five cross-fitted rotations. Post-selection test macro-F1 is $0.842\pm0.004$ (mean $\pm$ s.d.\ over rotations), which is the primary estimate of what the procedure delivers, and ECE after post-hoc calibration is $0.0043\pm0.0009$; the post-hoc temperature is $T^{\star}=0.90\pm0.04$ and is below unity in every rotation. Selection is sensitive to seed identity---four distinct triads are chosen across the five rotations---but stable at the family level: MentalRoBERTa is a member in every rotation, RoBERTa-base in four, MentalBERT in three, and Bio-ClinicalBERT in one. On the full validation set, the selected triad is MentalBERT\,s43 $+$ MentalRoBERTa\,s43 $+$ RoBERTa-base\,s42 (validation macro-F1 $0.868$, $\mathrm{CI}(S)=0.095$, three families), which is also the triad chosen in two of the five rotations; it is reported as the deployed instance in the remaining tables.

Because $\mathrm{CI}(S)$ enters the rule only as a tiebreaker, its influence depends on what counts as comparable validation performance, a tolerance the protocol leaves implicit. Table~\ref{tab:ci_ablation} therefore repeats the five cross-fitted rotations with and without the tiebreaker under four tolerances. At exact equality, no two of the $216$ candidates ever tie, so the tiebreaker never activates and every triad reported in this paper was selected by validation macro-F1 alone. Under a tolerance of $0.001$ the selection changes in two rotations and the mean test macro-F1 rises by $0.0005$; at $0.002$ and $0.005$ it changes in three and four rotations and falls by $0.0007$ and $0.0032$. The index therefore does not improve ensemble selection at any tolerance examined, and no such claim is made; its demonstrated role is the predictive one of Section~\ref{subsec:pairs}.

\begin{table}[!htbp]
\centering
\caption{Selection ablation: the five cross-fitted rotations repeated with and without $\mathrm{CI}(S)$ as a tiebreaker, for four definitions of ``comparable'' validation macro-F1. Test macro-F1 is the mean over the five rotations.}
\label{tab:ci_ablation}
\renewcommand{\arraystretch}{1.15}
\footnotesize
\begin{adjustbox}{max width=\textwidth}
\begin{tabular}{@{}lcccc@{}}
\toprule
Tolerance & Mean candidates tied & Selection changed & Test macro-F1, without CI & with CI\\
\midrule
$0$ (exact)   & 1.0  & 0/5 & 0.8420 & 0.8420\\
$0.001$       & 1.8  & 2/5 & 0.8420 & 0.8425\\
$0.002$       & 3.0  & 3/5 & 0.8420 & 0.8413\\
$0.005$       & 11.4 & 4/5 & 0.8420 & 0.8388\\
\bottomrule
\end{tabular}
\end{adjustbox}
\end{table}

\begin{table}[!htbp]
\centering
\caption{Cross-fitted selection: in rotation $k$, member temperatures are fitted on validation fold $k$, the triad is selected on the three remaining folds, and $T^{\star}$ is fitted on fold $k{+}1$; each resulting system is scored once on the test set. Test ECE is given before and after the post-hoc temperature. BC: Bio-ClinicalBERT; MB: MentalBERT; MR: MentalRoBERTa; RB: RoBERTa-base.}
\label{tab:crossfit}
\renewcommand{\arraystretch}{1.15}
\footnotesize
\begin{adjustbox}{max width=\textwidth}
\begin{tabular}{@{}clccc@{}}
\toprule
$k$ & Selected triad & $T^{\star}$ & Test macro-F1 & Test ECE (before $\to$ after $T^{\star}$)\\
\midrule
0 & MB\,s43 + MR\,s43 + RB\,s42 & 0.859 & 0.8442 & 0.0079 $\to$ 0.0040\\
1 & MB\,s42 + MR\,s43 + RB\,s43 & 0.929 & 0.8472 & 0.0059 $\to$ 0.0038\\
2 & MR\,s42 + MR\,s44 + RB\,s42 & 0.901 & 0.8363 & 0.0085 $\to$ 0.0046\\
3 & BC\,s44 + MR\,s42 + MR\,s44 & 0.878 & 0.8388 & 0.0085 $\to$ 0.0057\\
4 & MB\,s43 + MR\,s43 + RB\,s42 & 0.944 & 0.8437 & 0.0046 $\to$ 0.0035\\
\midrule
\multicolumn{2}{l}{Mean $\pm$ s.d.} & $0.902\pm0.035$ & $0.8420\pm0.0044$ & $0.0071\pm0.0017 \to 0.0043\pm0.0009$\\
\bottomrule
\end{tabular}
\end{adjustbox}
\end{table}

Table~\ref{tab:family_triads} removes the seed dimension by evaluating every family-level triad over all $27$ seed instantiations. The four family triads lie within $0.004$ of one another ($0.836$--$0.840$), each with a seed standard deviation of $0.004$--$0.007$; the triad without Bio-ClinicalBERT has the highest mean validation and test macro-F1. Every family triad exceeds the validation-selected best single encoder ($0.822$) by at least $0.014$ on average.

\begin{table}[!htbp]
\centering
\caption{Family-level triads: mean $\pm$ s.d.\ of macro-F1 over the $27$ seed instantiations of each triad (one seed per family), and mean test ECE after post-hoc calibration.}
\label{tab:family_triads}
\renewcommand{\arraystretch}{1.15}
\footnotesize
\begin{adjustbox}{max width=\textwidth}
\begin{tabular}{@{}lccc@{}}
\toprule
Family triad & Val.\ macro-F1 & Test macro-F1 & Test ECE\\
\midrule
MentalBERT + MentalRoBERTa + RoBERTa-base     & $0.863\pm0.002$ & $0.840\pm0.006$ & 0.0035\\
Bio-ClinicalBERT + MentalRoBERTa + RoBERTa-base & $0.860\pm0.003$ & $0.839\pm0.007$ & 0.0035\\
Bio-ClinicalBERT + MentalBERT + MentalRoBERTa & $0.860\pm0.003$ & $0.839\pm0.004$ & 0.0036\\
Bio-ClinicalBERT + MentalBERT + RoBERTa-base  & $0.860\pm0.004$ & $0.836\pm0.006$ & 0.0030\\
\bottomrule
\end{tabular}
\end{adjustbox}
\end{table}

\subsection{$N$-way ensembles}
\label{subsec:nway}

Table~\ref{tab:nway} compares the headline triad with the validation-selected best single encoder, homogeneous three-seed ensembles, and the twelve-model ensemble; Fig.~\ref{fig:roc} shows ROC curves and Fig.~\ref{fig:conf} shows confusion matrices. The heterogeneous triad ($0.8436$) exceeds the homogeneous three-seed MentalRoBERTa ensemble ($0.8379$), which is built from the strongest family, and the homogeneous RoBERTa-base ensemble ($0.8299$). The twelve-model uniform ensemble reaches $0.8478$, i.e.\ $0.004$ above the triad at four times its inference cost, with overlapping intervals. Uniform averaging is the pre-specified fusion rule; learned per-class weights (Eq.~\eqref{eq:ensemble}) were evaluated against it on validation data within the cross-fitted rotations. Fitted on the selection folds, the weights raise in-sample macro-F1 by $+0.005$ on average, but on the held-out validation fold they exceed uniform averaging in only two of five rotations (mean $\Delta=+0.003$, s.d.\ $0.008$, one rotation contributing $+0.016$ and the others between $-0.002$ and $+0.003$). The eight free parameters therefore do not improve macro-F1 consistently out of sample within validation, and uniform averaging is retained on that basis. For completeness, the test macro-F1 of the weighted and uniform versions of the same rotations differ by $-0.0007$ on average, and weights fitted on the full validation set give $0.8437$ versus $0.8436$ for the deployed triad and $0.8453$ versus $0.8478$ for the twelve-model ensemble; these test figures played no part in the decision.

\begin{table}[!htbp]
\centering
\caption{$N$-way ensembles on the lineage-disjoint test set. All ensembles are uniform means of calibrated member probabilities followed by a post-hoc temperature fitted on the validation set; ECE is given before and after $T^{\star}$. The diverse triad is MentalBERT\,s43 + MentalRoBERTa\,s43 + RoBERTa-base\,s42.}
\label{tab:nway}
\renewcommand{\arraystretch}{1.15}
\footnotesize
\setlength{\tabcolsep}{4pt}
\begin{adjustbox}{max width=\textwidth}
\begin{tabular}{@{}lcccc@{}}
\toprule
System & macro-F1 & MCC & $T^{\star}$ & ECE (before $\to$ after)\\
\midrule
Best single by validation (MentalRoBERTa\,s44) & 0.8225 [0.8012, 0.8429] & 0.7648 & 1.00 & 0.0066 $\to$ 0.0066\\
Homogeneous $3\times$RoBERTa-base                & 0.8299 [0.8063, 0.8513] & 0.7667 & 0.94 & 0.0076 $\to$ 0.0070\\
Homogeneous $3\times$MentalRoBERTa              & 0.8379 [0.8164, 0.8567] & 0.7752 & 0.95 & 0.0051 $\to$ 0.0034\\
\textbf{Diverse triad} & \textbf{0.8436 [0.8231, 0.8625]} & \textbf{0.7791} & 0.89 & 0.0063 $\to$ \textbf{0.0033}\\
All twelve, uniform                              & 0.8478 [0.8278, 0.8667] & 0.7852 & 0.83 & 0.0120 $\to$ 0.0028\\
All twelve, learned per-class weights           & 0.8453 & --- & --- & ---\\
\bottomrule
\end{tabular}
\end{adjustbox}
\end{table}

\begin{figure*}[!htbp]
\centering
\includegraphics[width=\textwidth]{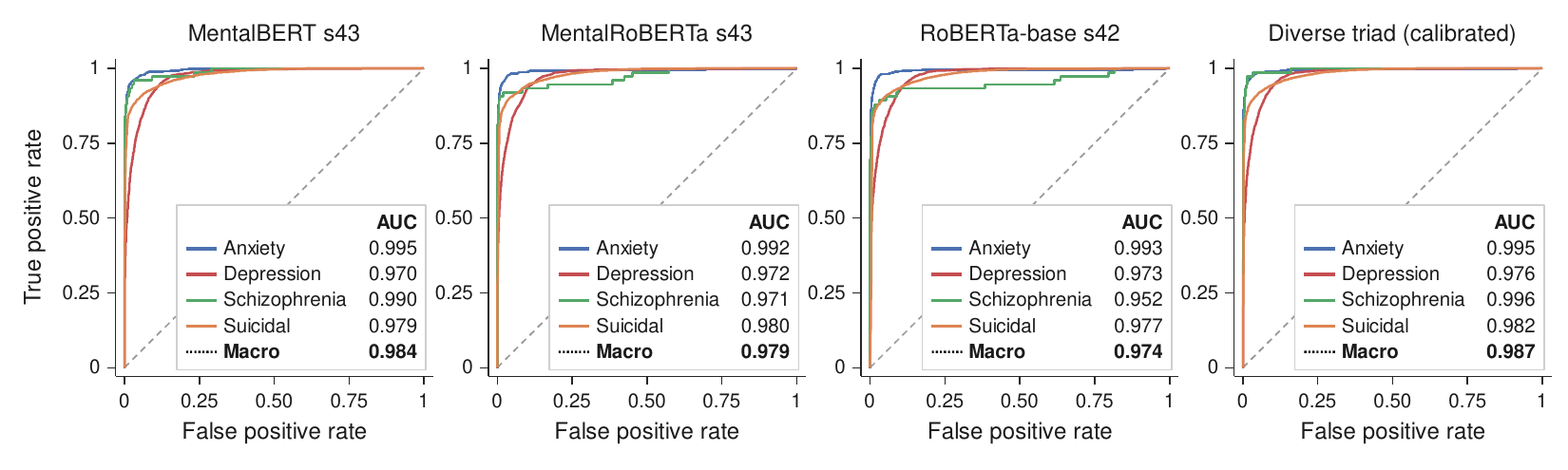}
\caption{One-versus-rest ROC curves of the three headline members and of the calibrated triad on the lineage-disjoint test set.}
\label{fig:roc}
\end{figure*}

\begin{figure*}[!htbp]
\centering
\includegraphics[width=\textwidth]{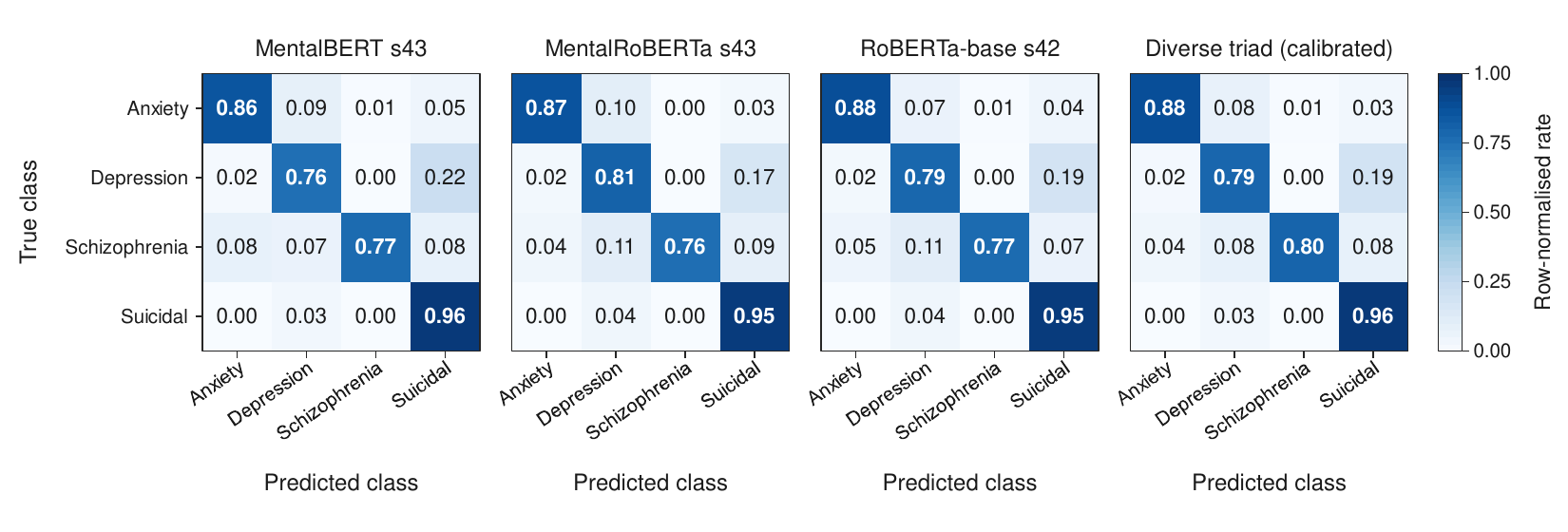}
\caption{Row-normalized confusion matrices of the three headline members and of the calibrated triad. The dominant residual confusion is depression$\to$suicidal ideation ($17$--$22\%$ of depression posts for the members, $19\%$ for the triad); the triad attains the highest recall on schizophrenia ($0.80$ versus $0.76$--$0.77$).}
\label{fig:conf}
\end{figure*}

\subsection{Post-hoc ensemble calibration}
\label{subsec:posthoc_results}

The post-hoc temperature fitted on the validation set for the headline triad is $T^{\star}=0.886$; its members' temperatures are $1.33$, $1.39$, and $1.45$. Applying $T^{\star}$ leaves every arg-max decision unchanged and reduces the test ECE from $0.0063$ $[0.0057,0.0096]$ to $0.0033$ $[0.0027,0.0064]$ ($47.6\%$; paired bootstrap interval of the reduction $[-0.0058,-0.0002]$). Fig.~\ref{fig:reliability} shows the reliability diagrams, and Table~\ref{tab:ece_robustness} shows that the reduction is not an artifact of the $15$-bin estimator: it is reproduced with $10$, $20$, and $30$ equal-width bins, with equal-mass bins, with the class-wise variant, and with the binning-free kernel estimator; the negative log-likelihood also decreases, and the Brier score, which is dominated by its refinement component, is unchanged to four decimals. The same direction holds for every ensemble in Table~\ref{tab:nway} and every rotation in Table~\ref{tab:crossfit}: the average of individually calibrated members is under-confident in domain, and a single sharpening temperature corrects it.

\begin{figure*}[!htbp]
\centering
\includegraphics[width=0.85\textwidth]{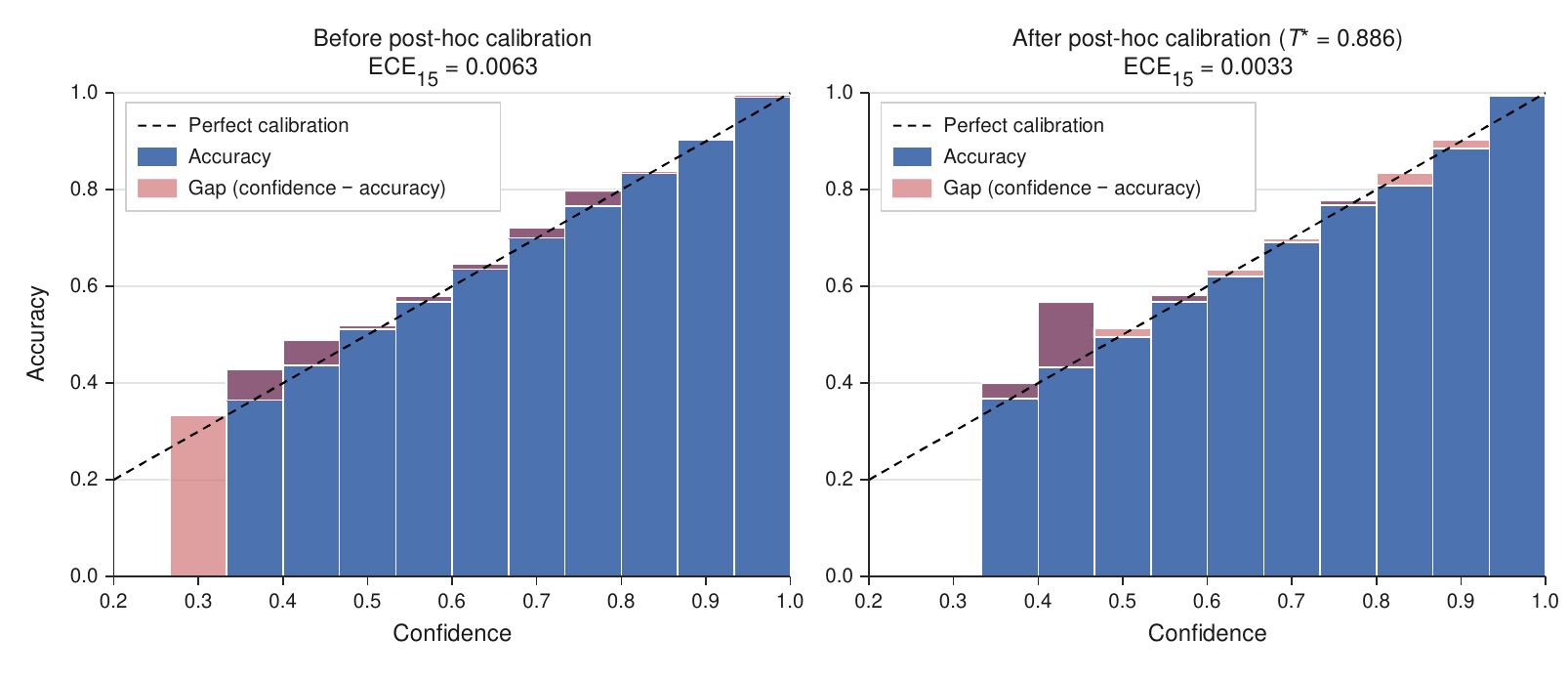}
\caption{Reliability diagrams of the headline triad before (left) and after (right) the post-hoc temperature, $15$ equal-width bins. The gap between confidence and accuracy is reduced without any change in the predicted labels.}
\label{fig:reliability}
\end{figure*}

\begin{table}[!htbp]
\centering
\caption{Sensitivity of the calibration result to the estimator: headline triad before and after the post-hoc temperature.}
\label{tab:ece_robustness}
\renewcommand{\arraystretch}{1.15}
\footnotesize
\begin{adjustbox}{max width=\textwidth}
\begin{tabular}{@{}lcc@{}}
\toprule
Estimator & Before $T^{\star}$ & After $T^{\star}$\\
\midrule
ECE, 10 equal-width bins   & 0.0063 & 0.0028\\
ECE, 15 equal-width bins   & 0.0063 & 0.0033\\
ECE, 20 equal-width bins   & 0.0072 & 0.0034\\
ECE, 30 equal-width bins   & 0.0076 & 0.0045\\
ECE, 15 equal-mass bins    & 0.0066 & 0.0042\\
Class-wise ECE, 15 bins    & 0.0121 & 0.0106\\
Kernel (binning-free) ECE  & 0.0079 & 0.0045\\
Negative log-likelihood    & 0.1447 & 0.1436\\
Brier score                & 0.0833 & 0.0833\\
\bottomrule
\end{tabular}
\end{adjustbox}
\end{table}

\subsection{Head-to-head comparison with the best single encoder}
\label{subsec:head2head}

Table~\ref{tab:head2head} compares the calibrated headline triad with the single encoder that a practitioner would select on validation (MentalRoBERTa\,s44). The triad improves macro-F1 by $+0.021$ (paired bootstrap $[+0.007,+0.037]$; permutation $p=0.0035$), MCC by $+0.014$, and Brier score by $0.0056$ (paired interval $[-0.0073,-0.0039]$), and halves the ECE ($\Delta=-0.0033$, $[-0.0058,-0.0003]$). McNemar's test on per-record correctness gives $b/c=237/345$, $\chi^2=19.7$, $p=9.2\times10^{-6}$, and $p=4.6\times10^{-4}$ after Holm correction over the $67$ ensemble-versus-member tests performed in this study. The improvement is distributed across classes. Schizophrenia gains $+0.060$ $[+0.009,+0.117]$ and depression $+0.018$ $[+0.009,+0.026]$, while anxiety gains $+0.004$ $[-0.008,+0.018]$ and suicidal ideation $+0.002$ $[+0.001,+0.004]$. Because macro-F1 weights the four classes equally, the $75$-post schizophrenia class contributes $+0.015$ of the $+0.021$ total, or $71\%$ of the headline gain (bootstrap interval of that share $[27\%,91\%]$). The gain does not depend on that class alone: restricted to the three classes with at least $781$ test posts, the macro-F1 improvement is $+0.008$ $[+0.003,+0.014]$ and still excludes zero. Both figures should be read together, since the headline number is dominated by the class with the weakest statistical support and the least certain provenance (Section~\ref{subsec:threats}). The single encoder that is strongest on the test set (MentalRoBERTa\,s42, $0.838$) is not identifiable from validation data; against it the triad's advantage is $+0.006$ and not significant, which is the appropriate reading of how much of the ensemble gain is recoverable by selecting one model well.

\begin{table}[!htbp]
\centering
\caption{Calibrated headline triad versus the validation-selected best single encoder on the lineage-disjoint test set. Bracketed values are $95\%$ bootstrap percentile intervals ($B=2000$); the $\Delta$ column and the tests below are paired over the same resamples. Differences are computed before rounding. Because macro-F1 weights the four classes equally, the $75$-post schizophrenia class contributes $+0.015$ of the $+0.021$ macro-F1 difference; restricted to the three classes with at least $781$ test posts, the difference is $+0.008$ $[+0.003,+0.014]$.}
\label{tab:head2head}
\renewcommand{\arraystretch}{1.2}
\footnotesize
\setlength{\tabcolsep}{4pt}
\begin{adjustbox}{max width=\textwidth}
\begin{tabular}{@{}lccc@{}}
\toprule
 & Best single (MentalRoBERTa\,s44) & Diverse triad, calibrated & $\Delta$ (triad $-$ single)\\
\midrule
Macro-F1      & 0.8225 [0.8012, 0.8429] & \textbf{0.8436 [0.8231, 0.8625]} & $+0.0211$ [$+0.0071$, $+0.0366$]\\
MCC           & 0.7648 [0.7534, 0.7754] & \textbf{0.7791 [0.7675, 0.7898]} & $+0.0143$ [$+0.0067$, $+0.0221$]\\
ECE           & 0.0066 [0.0052, 0.0097] & \textbf{0.0033 [0.0027, 0.0064]} & $-0.0033$ [$-0.0058$, $-0.0003$]\\
Brier         & 0.0889 [0.0850, 0.0928] & \textbf{0.0833 [0.0796, 0.0869]} & $-0.0056$ [$-0.0073$, $-0.0039$]\\
Accuracy      & 0.9362 & \textbf{0.9410} & $+0.0048$ [$+0.0027$, $+0.0070$]\\
Balanced acc. & 0.8474 & \textbf{0.8594} & $+0.0121$ [$-0.0045$, $+0.0299$]\\
\midrule
F1 Anxiety       & 0.879 [0.862, 0.896] & \textbf{0.884 [0.866, 0.900]} & $+0.004$ [$-0.008$, $+0.018$]\\
F1 Depression    & 0.740 [0.728, 0.753] & \textbf{0.758 [0.745, 0.770]} & $+0.018$ [$+0.009$, $+0.026$]\\
F1 Schizophrenia & 0.704 [0.618, 0.785] & \textbf{0.764 [0.686, 0.836]} & $+0.060$ [$+0.009$, $+0.117$]\\
F1 Suicidal      & 0.966 [0.964, 0.968] & \textbf{0.969 [0.967, 0.970]} & $+0.002$ [$+0.001$, $+0.004$]\\
\midrule
\multicolumn{4}{@{}l}{Macro-F1 difference, permutation test: $p=0.0035$}\\
\multicolumn{4}{@{}l}{McNemar on per-record correctness: $b/c=237/345$; $\chi^{2}=19.67$; $p=9.2\times10^{-6}$; $p=4.6\times10^{-4}$ after Holm correction over $67$ tests}\\
\bottomrule
\end{tabular}
\end{adjustbox}
\end{table}
\subsection{Lexical baseline and artifact controls}
\label{subsec:controls}

A TF--IDF logistic-regression classifier trained on the lineage-disjoint training set reaches macro-F1 $0.748$ $[0.727,0.768]$ on the test set (per class $0.79$, $0.57$, $0.68$, $0.95$), i.e.\ $0.096$ below the triad, with the largest deficit on depression ($0.57$ versus $0.76$). Trained on the $104{,}436$ original training posts alone, without augmentation, the same classifier reaches $0.489$, with schizophrenia F1 falling to $0.06$: augmentation within the training partition is what makes the smallest classes learnable, and it is confined to that partition. Table~\ref{tab:artifacts} reports the masking controls for the triad. Masking Reddit meta-vocabulary, present in $17.7\%$ of test posts, changes macro-F1 by $+0.001$, so the tested platform vocabulary carries no usable signal. This control covers an explicit word list only; broader source- or style-level artifacts, which cannot be masked lexically, are not excluded by it. Additionally, masking explicit condition names, present in $45.4\%$ of test posts, lowers macro-F1 to $0.753$, with the loss concentrated on schizophrenia ($0.764 \to 0.537$) and anxiety ($0.884 \to 0.800$), smaller on depression ($0.758 \to 0.711$) and negligible on suicidal ideation ($0.969 \to 0.965$). Because masking produces inputs unlike any seen in training, the in-distribution figure is the more informative one: on the $54.6\%$ of test posts that never name a condition, the triad reaches macro-F1 $0.792$.

\begin{table}[!htbp]
\centering
\caption{Artifact controls for the headline triad on the lineage-disjoint test set.}
\label{tab:artifacts}
\renewcommand{\arraystretch}{1.15}
\footnotesize
\begin{adjustbox}{max width=\textwidth}
\begin{tabular}{@{}lccccc@{}}
\toprule
Test-set variant & macro-F1 & Anx & Dep & Scz & Sui\\
\midrule
Original                                              & 0.844 & 0.884 & 0.758 & 0.764 & 0.969\\
Reddit meta-vocabulary masked ($17.7\%$ of posts)     & 0.845 & 0.884 & 0.756 & 0.772 & 0.968\\
$+$ condition names masked ($45.4\%$ of posts)        & 0.753 & 0.800 & 0.711 & 0.537 & 0.965\\
Natural subset: posts with no condition name ($54.6\%$) & 0.792 & \multicolumn{4}{c}{---}\\
\midrule
TF--IDF + logistic regression (full test set)         & 0.748 & 0.790 & 0.574 & 0.682 & 0.945\\
\bottomrule
\end{tabular}
\end{adjustbox}
\end{table}

\subsection{Zero-shot transfer to a same-platform external corpus}
\label{subsec:external}

Table~\ref{tab:external} reports zero-shot transfer to the author-disjoint SDCNL test half. The triad reaches two-class macro-F1 $0.723$ $[0.693,0.750]$. Against the member selected on the external validation half (MentalBERT\,s43, $0.699$ on the test half) the gain is $+0.024$ $[+0.002,+0.047]$ (McNemar $b/c=44/66$, $p=0.045$); against the member that happens to be strongest on the external test half (MentalRoBERTa\,s43, $0.714$), which is not identifiable from validation data, it is $+0.008$ $[-0.012,+0.029]$ and not significant. The four-class models assign a class outside \{depression, suicidal ideation\} to only $3$--$5\%$ of external posts. Calibration behaves differently under transfer: every temperature fitted on the external validation half exceeds unity ($1.4$--$2.4$), i.e.\ the models and their ensemble are over-confident out of domain, and the post-hoc step fitted on $892$ external posts halves the triad's ECE ($0.066\to0.033$). CI computed on the external validation half shows no correlation with gain on the external test half ($r=0.05$ over $12$ pair--class observations), but all gains lie within $\pm0.03$ and all CI values within $0.18$--$0.29$, so the test is inconclusive at this sample size rather than negative.

\begin{table}[!htbp]
\centering
\caption{Zero-shot transfer to SDCNL (external test half, $N=916$, author-disjoint from the external validation half; $83$ near-duplicates of the main corpus removed). Two-class macro-F1 with $95\%$ bootstrap intervals; ECE before and after a post-hoc temperature fitted on the external validation half.}
\label{tab:external}
\renewcommand{\arraystretch}{1.15}
\footnotesize
\begin{adjustbox}{max width=\textwidth}
\begin{tabular}{@{}lccccc@{}}
\toprule
System & macro-F1 & F1 depression & F1 suicidal & ECE (before $\to$ after) & $T^{\star}_{\text{ext}}$\\
\midrule
MentalBERT\,s43        & 0.699 [0.668, 0.727] & 0.695 & 0.702 & 0.078 $\to$ 0.027 & 1.76\\
MentalRoBERTa\,s43     & 0.714 [0.686, 0.745] & 0.730 & 0.699 & 0.070 $\to$ 0.051 & 1.79\\
RoBERTa-base\,s42      & 0.693 [0.664, 0.723] & 0.704 & 0.682 & 0.105 $\to$ 0.054 & 2.40\\
Bio-ClinicalBERT\,s42  & 0.678 [0.648, 0.707] & 0.683 & 0.673 & 0.079 $\to$ 0.041 & 2.02\\
\textbf{Diverse triad} & \textbf{0.723 [0.693, 0.750]} & 0.727 & 0.718 & 0.066 $\to$ \textbf{0.033} & 1.50\\
All four               & 0.718 [0.689, 0.749] & 0.723 & 0.714 & 0.053 $\to$ 0.037 & 1.37\\
\bottomrule
\end{tabular}
\end{adjustbox}
\end{table}
\end{revblock}

\FloatBarrier

\section{Attribution-Diversity Analysis and Counterfactual Intervention}
\label{sec:xai}

\begin{revblock}
The question addressed in this section is which property of the members drives the fusion benefit. Integrated Gradients (Eq.~\eqref{eq:ig}) were computed for the three deployed checkpoints of the headline triad on $1{,}575$ test posts (all $75$ schizophrenia posts and $500$ per other class), and attributions were aligned at word level as described in Section~\ref{subsec:ig}. A post is an all-correct case if the ensemble and all three members classify it correctly, a fusion-rescued case if the ensemble is correct and at least one member is wrong, and an ensemble-failure case if the ensemble is wrong. The sample contains $1{,}250$ all-correct cases ($79.4\%$), $122$ fusion-rescued cases ($7.7\%$; $101$ with exactly one incorrect member and $21$ with two), and $203$ ensemble-failure cases ($12.9\%$).

\subsection{Attribution diversity is higher on fusion-rescued cases}
\label{subsec:diversity}

For each post, the attribution-diversity score is $1$ minus the mean pairwise cosine similarity of the three word-level attribution vectors. Table~\ref{tab:diversity} and Fig.~\ref{fig:div_scatter}a give the result. All-correct cases have a mean diversity of $0.401$ and fusion-rescued cases $0.904$ ($\Delta=+0.503$, permutation $p=0.0001$ with $10{,}000$ shuffles); ensemble-failure cases lie in between ($0.646$). The effect is present within every class ($\Delta$ between $+0.36$ and $+0.63$, all $p<0.001$), so it is not a by-product of the classes' different base rates of diversity.

Two controls address confounds specific to attribution methods. First, IG attributes a model's own predicted class; a member that predicts a different class therefore attributes a different logit, which inflates diversity on rescued cases by construction. Restricting the comparison to member pairs that predict the same class removes this effect: same-target diversity is $0.401$ on all-correct cases and $0.610$ on rescued cases ($\Delta=+0.21$, $p=0.0001$; $n=115$ rescued cases with at least one same-target pair), and $0.614$ on the $101$ rescued cases with exactly one incorrect member, whose two correct members always share the target. Second, position-wise comparison of token-level vectors from different tokenizers, without word alignment, gives a baseline diversity of $0.679$ on all-correct cases, i.e.\ it manufactures disagreement from tokenization alone; word-level alignment reduces this to $0.401$ while preserving the between-group difference. A top-20-word Jaccard distance shows no effect ($\Delta=+0.01$, $p=0.33$): the members largely agree on which words matter and differ in the sign and magnitude they assign to them.

Per class (Table~\ref{tab:diversity}), the rescue rate is highest for depression ($12.4\%$), the class with the highest mean diversity ($0.551$), followed by schizophrenia ($9.3\%$), anxiety ($7.8\%$), and suicidal ideation ($2.8\%$). The rescue rate and the per-class fusion gain of Table~\ref{tab:head2head} are both largest on the two classes where the members disagree most, but their orderings differ: depression has the highest rescue rate, whereas schizophrenia shows the largest F1 gain.

\begin{table}[!htbp]
\centering
\caption{Word-level inter-model attribution diversity ($1-$ mean pairwise cosine) on the $1{,}575$ analysed test posts. Permutation $p$-values use $10{,}000$ label shuffles.}
\label{tab:diversity}
\renewcommand{\arraystretch}{1.15}
\footnotesize
\setlength{\tabcolsep}{4pt}
\begin{adjustbox}{max width=\textwidth}
\begin{tabular}{@{}lcccc@{}}
\toprule
 & All-correct & Fusion-rescued & $\Delta$ & $p$\\
\midrule
All member pairs                 & 0.401 ($n=1{,}250$) & 0.904 ($n=122$) & $+0.503$ & 0.0001\\
Same-target pairs only           & 0.401 ($n=1{,}250$) & 0.610 ($n=115$) & $+0.209$ & 0.0001\\
Position-wise, no word alignment & 0.679 & 0.946 & $+0.268$ & 0.0001\\
Top-20-word Jaccard distance     & 0.641 & 0.652 & $+0.011$ & 0.33\\
\midrule
\multicolumn{5}{@{}l}{Within class, all pairs ($\Delta$, $n$ rescued): Anxiety $+0.61$ (39); Depression $+0.36$ (62);}\\
\multicolumn{5}{@{}l}{\quad Schizophrenia $+0.63$ (7); Suicidal $+0.55$ (14); all $p<0.001$}\\
\midrule
Class & Test posts & Rescue rate & Failure rate & Mean diversity\\
\midrule
Anxiety       & 500 & 7.8\%  & 10.6\% & 0.462\\
Depression    & 500 & 12.4\% & 21.6\% & 0.551\\
Schizophrenia & 75  & 9.3\%  & 20.0\% & 0.454\\
Suicidal      & 500 & 2.8\%  & 5.4\%  & 0.405\\
\bottomrule
\end{tabular}
\end{adjustbox}
\end{table}

\subsection{Counterfactual word removal}
\label{subsec:counterfactual}

The association above does not establish that the divergent evidence causes the rescue. An instance-level intervention was therefore performed. For each of the $98$ rescued cases with exactly one incorrect member and at least $15$ aligned words, three sets of five words were removed from the post, and the three members were re-scored: (i) the five divergent words with the largest difference between the mean attribution of the two correct members and the attribution of the incorrect member, (ii) the five consensus words with the largest minimum attribution across all three members (excluding the divergent set), and (iii) five words drawn at random (three independent draws). The same procedure was applied to $282$ all-correct cases as a reference, with the divergent set defined by the largest spread of attribution across members.

Fig.~\ref{fig:div_scatter}b and Table~\ref{tab:counterfactual} report the fraction of cases on which the ensemble decision remains correct. On rescued cases, removing five random words preserves the rescue $88.4\%$ of the time and removing five consensus words $85.7\%$, whereas removing the five divergent words preserves it only $41.8\%$ of the time (paired sign-permutation test against the random condition: $\Delta=0.47$, $p=0.0001$; consensus versus random: $p=0.22$). When a rescue is destroyed by divergent-word removal, the ensemble adopts the class predicted by the originally incorrect member in $91\%$ of cases ($52$ of $57$). On all-correct cases, the same removal changes the decision far less often ($81.2\%$ preserved), so rescued decisions are specifically dependent on the words on which the members disagree. This shows that rescued decisions are markedly more sensitive to removal of the divergent words than to matched consensus or random removals. It does not isolate attribution divergence as the cause: the selected words may also be the most informative ones, and their removal changes the surrounding context.

\begin{table}[!htbp]
\centering
\caption{Counterfactual word removal: fraction of cases on which the ensemble decision remains correct after removing five words selected by each rule. Random is the mean of three draws.}
\label{tab:counterfactual}
\renewcommand{\arraystretch}{1.15}
\footnotesize
\begin{adjustbox}{max width=\textwidth}
\begin{tabular}{@{}lcc@{}}
\toprule
Words removed & Fusion-rescued ($n=98$) & All-correct ($n=282$)\\
\midrule
None                & 1.000 & 1.000\\
Five random         & 0.884 & 0.983\\
Five consensus      & 0.857 & 0.979\\
Five divergent      & \textbf{0.418} & 0.812\\
\midrule
Divergent vs.\ random, paired $\Delta$ ($p$) & $0.47$ ($0.0001$) & $0.17$ ($0.0001$)\\
Consensus vs.\ random, paired $\Delta$ ($p$) & $0.03$ ($0.22$) & $0.005$ ($0.36$)\\
\bottomrule
\end{tabular}
\end{adjustbox}
\end{table}

\begin{figure}[!htbp]
\centering
\includegraphics[width=\columnwidth]{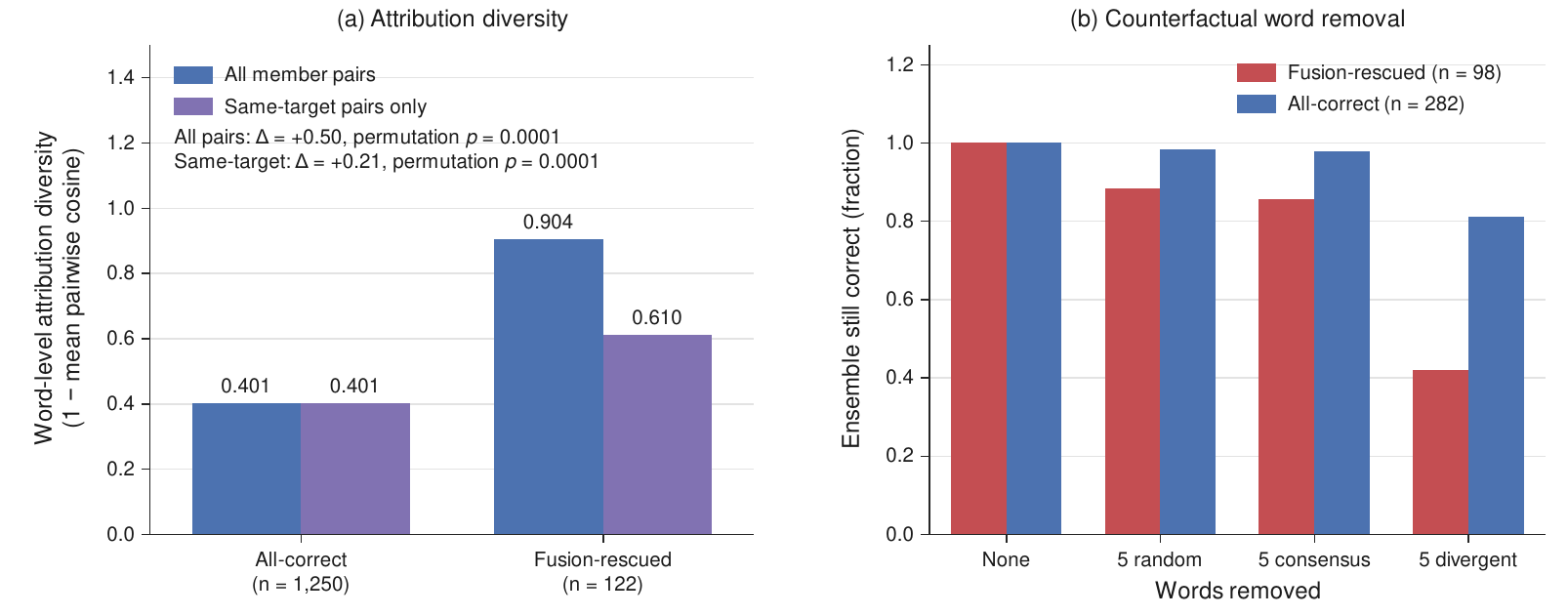}
\caption{(a) Word-level attribution diversity on all-correct versus fusion-rescued cases, for all member pairs and for same-target pairs only. (b) Fraction of decisions preserved after removing five words by rule; rescued decisions depend on the divergent words.}
\label{fig:div_scatter}
\end{figure}

\subsection{Interpretation}
\label{subsec:xai_interp}

Fusion rescue is concentrated on posts for which the members weight the same words differently, and it is the words on which the correct members diverge from the incorrect ones that carry the rescue. Together with Section~\ref{subsec:pairs}, this supports a consistent account: the ensemble helps on the confusable classes because its members extract partly different evidence from the same text, and CI, which counts class-specific examples that only one member classifies correctly, is a validation-time proxy for exactly this divergence. The intervention operates on inputs; a model-side intervention that enforces one member's attribution pattern on another through attribution regularization~\citep{rieger2020interpretations} would test sufficiency and remains future work.
\end{revblock}

\FloatBarrier
\section{Discussion}
\label{sec:discussion}

\begin{revblock}
\subsection{What a leakage audit must cover}
\label{subsec:disc_leakage}

The two-layer audit of Section~\ref{subsec:audit} has a general lesson for corpora assembled from public collections: near-duplicate detection answers the question ``is this evaluation record close to a training record?'', whereas the relevant question is ``does this evaluation record share provenance with a training record?''. When augmentation, re-posting, or cross-posting precedes splitting, the second question has a different answer from the first. On the random split, the lexical audit at Jaccard $\geq0.8$ removed $29\%$ of the test records and left $30\%$ of the remainder lineage-linked; on the same corpus, the lineage-linked fraction was $100\%$ for a class with $500$ original posts and $34{,}500$ derivatives. The corresponding consequence for reported performance is large: the per-class F1 of the schizophrenia class, which is close to $1$ when derivatives of its $500$ originals are distributed across splits, is $0.62$--$0.77$ with intervals of $\pm0.08$ when they are not. 

\subsection{Encoder families}
\label{subsec:disc_families}

Bio-ClinicalBERT was pretrained on MIMIC-III clinical notes, which are written in a clinical register, whereas the target text is self-authored and affective. The mental-health-specialized families exceed Bio-ClinicalBERT by $0.02$--$0.03$ seed-averaged macro-F1, consistently across seeds, which is consistent with a register-mismatch account of specialist pretraining. General-domain RoBERTa-base, however, matches MentalBERT on average and exhibits the largest seed variance of any family, so the ordering among the three non-clinical families is not resolvable at three seeds and should not be over-interpreted. Family membership matters for fusion in a different way: the cross-fitted protocol selects MentalRoBERTa in every rotation and pairs it with a RoBERTa-base or MentalBERT member in most, and family-level triads that combine two backbone families with two pretraining corpora lie within $0.004$ of one another.

\subsection{Uniform versus learned weights}
\label{subsec:disc_weights}

The learned per-class weighting (Eq.~\eqref{eq:ensemble}) has $K(M-1)$ free parameters under the per-class simplex constraint: eight for a triad and $44$ for twelve members. Within the cross-fitted rotations, it improves in-sample macro-F1 on the selection folds but not consistently on the held-out validation fold (two of five rotations). The validation set is large ($22{,}379$ posts) but the two smallest classes contribute $857$ of them, and the weights are fitted on macro-F1, in which those classes carry half the weight; the parameters therefore fit noise in the minority classes. Uniform averaging, the pre-specified rule, is retained on validation evidence alone.

\subsection{Direction of the post-hoc temperature}
\label{subsec:disc_temperature}

Every member temperature exceeds unity ($1.32$--$1.52$): fine-tuned encoders are over-confident, and member-wise scaling softens them. The uniform mean of softened distributions is then under-confident in domain---the winning-class probability is lower than the empirical accuracy---and $T^{\star}<1$ in every rotation, ensemble size, and family triad examined ($0.83$--$0.95$). Under transfer to SDCNL, the sign reverses: the members and their ensemble are over-confident on the external corpus and $T^{\star}_{\text{ext}}$ lies between $1.4$ and $2.4$. The two-stage procedure is what transfers; the value and sign of $T^{\star}$ do not, and a post-hoc temperature should be refitted on a small validation slice of any new domain, where $892$ posts were sufficient to halve the ECE.

\subsection{Threats to validity}
\label{subsec:threats}

\paragraph{Data source and labels}
The corpus is drawn from self-selected online communities, and the labels are proxies for a self-declared condition; the results concern patient-authored social-media language and not clinical diagnosis~\citep{harrigian2021state,chancellor2020methods,ernala2019methodological}. The lexical baseline and the masking controls show that the task is not lexically trivial and does not depend on platform artifacts, but explicit self-labeling contributes to the anxiety and schizophrenia classes, and the figure of $0.792$ on posts that never name a condition is the appropriate estimate of performance without it. Manual inspection of the schizophrenia originals also found a small number of records written in a third-person clinical register rather than as forum posts; their provenance could not be traced in the merged file.

\paragraph{Split integrity}
The lineage-disjoint split removes contamination through augmentation lineage and leaves $0.4\%$ lexical near-duplicates among original posts, which are masked. It does not resolve every form of overlap: author-, thread-, and temporal-level contamination among original posts cannot be audited because the corresponding metadata were not retained. Edited re-posts among originals are covered by the semantic audit down to a cosine of $0.90$ ($0.8\%$ of evaluation posts, without effect on the results), but not below it. The external evaluation, which is author-disjoint by construction, partly compensates for the first of these. The parent of each derivative was recovered by similarity rather than from records; a misassignment could place a derivative across partitions only if it were more similar to another original than to its own parent, and the residual lexical audit bounds the effect.

\paragraph{Small-class statistics}
The schizophrenia class has $75$ validation and $75$ test posts. Its F1 intervals are $\pm0.08$ wide, its per-pair fusion gains are dominated by sampling noise, and CI is uninformative for it; any conclusion specific to this class rests on weak evidence. Because macro-F1 weights classes equally, this class also carries $71\%$ of the headline improvement over the best single encoder; the improvement restricted to the three larger classes is $+0.008$ $[+0.003,+0.014]$.

\paragraph{Selection stability}
Ensemble selection is stable at the family level but not at the seed level: four different triads are chosen across five validation rotations. The headline system is therefore reported together with the post-selection spread ($0.842\pm0.004$), and the family-level results, and the single-encoder comparison uses the encoder that validation data would select rather than the one that happens to be strongest on the test set.

\paragraph{Scope of the CI}
CI is associated with fusion gain within the two largest and most confusable classes and not within anxiety; the pooled correlation across classes is uninformative, and leave-one-encoder-out prediction has low out-of-sample accuracy. CI is presented as a class-conditional diagnostic; the ablation of Table~\ref{tab:ci_ablation} shows no selection benefit at any tolerance examined, and the external test of its transfer is inconclusive at $916$ posts. The $66$ pairs are formed from $12$ encoders, so pair-level statistics are not independent; all CI intervals reported here are encoder-aware, and under that correction the pooled slope across all four classes is not significant.

\paragraph{External transfer}
The external corpus is drawn from the same platform as the main corpus, covers two of the four classes and one language, and is used zero-shot; it tests transfer of the trained system and of the calibration procedure to an independently collected, author-disjoint sample, not cross-platform or clinical validity of the four-class task. Evaluation on clinical interviews, cross-platform settings, or non-English text remains open.

\paragraph{Attribution analysis}
Integrated Gradients is one attribution method, with a zero baseline and $20$ steps; the word-level alignment and same-target control address the two identified confounds, and the counterfactual intervention establishes the necessity of the divergent words for rescue on the analyzed cases. 
\end{revblock}

\section{Conclusion}
\label{sec:conclusion}

\begin{revblock}
A leakage-aware, calibration-aware decision-level fusion framework was proposed for four-class psychiatric text classification from user-generated text and evaluated under a protocol designed to withstand the failure modes of proxy-labeled corpora. A two-layer split-integrity audit showed that lexical near-duplicate detection, although it removed $29\%$ of a random test split, left $30\%$ of the retained records lineage-linked to training posts through augmentation lineage; a lineage-disjoint split was constructed. Twelve encoders across four families and three seeds were fine-tuned; every post-training decision, including the fusion rule, was made on validation folds disjoint from the fold used for the preceding step, and the test set was used for reporting only. On this basis, cross-fitted three-member ensembles reached a post-selection test macro-F1 of $0.842\pm0.004$, and the deployed triad of MentalBERT, MentalRoBERTa, and RoBERTa-base achieved macro-F1 $0.844$ $[0.823,0.863]$, MCC $0.779$, and ECE $0.0033$, exceeding the validation-selected best single encoder by $+0.021$ macro-F1 under paired-bootstrap, permutation, and multiplicity-corrected McNemar tests, and coming within $0.004$ of a twelve-model ensemble at one quarter of the cost. Two-stage calibration reduced the ensemble's ECE by $47.6\%$ under the $15$-bin estimator. Every other binned, equal-mass, class-wise and binning-free estimator examined moved in the same direction (reductions of $12$--$56\%$), while the negative log-likelihood decreased slightly and the Brier score was unchanged; the post-hoc correction sharpened the aggregate in domain and softened it under external transfer. The Complementarity Index, computed on validation and evaluated against test gain with dependence-aware statistics, was informative for the depression and suicidal-ideation classes and not for anxiety. The reported triads were selected by validation macro-F1 alone, and the index is therefore reported as a diagnostic rather than as a selection mechanism. Word-aligned Integrated-Gradients attributions on the deployed checkpoints were $2.3\times$ more diverse on fusion-rescued cases; the effect survived a same-target control, and a counterfactual intervention showed that removing the five words on which the correct members diverge from the incorrect one destroyed $58\%$ of rescues, compared with $12\%$ for random words. Future work includes model-side interventions on attribution patterns, external corpora that cover all four classes with author identifiers, and evaluation on clinical-interview text.
\end{revblock}

\section*{Acknowledgments}
The authors would like to thank Prince Sultan University for its support.

\section*{Declaration of Generative AI and AI-assisted technologies in the writing process}
During the preparation of this manuscript, the authors used Claude and ChatGPT to improve its content, clarity, and readability. The authors reviewed and edited all AI-assisted content and take full responsibility for the final published article.

\bibliographystyle{plainnat}
\bibliography{references}

\end{document}